\documentclass[10pt,twocolumn,letterpaper]{article}

\usepackage{cvpr}
\usepackage{times}
\usepackage{epsfig}
\usepackage{graphicx}
\usepackage{amsmath}
\usepackage{amssymb}
\usepackage{booktabs, multicol, multirow}

\usepackage{verbatim}
\usepackage{xfrac}
\usepackage{subfig}
\usepackage{mathtools}
\usepackage{times}
\usepackage{epsfig}
\usepackage{graphicx}
\usepackage{amsmath}
\usepackage{amssymb}
\usepackage{stfloats}
\usepackage{combelow}
\usepackage{flushend}
\usepackage{gensymb}

\newcommand\blfootnote[1]{%
  \begingroup
  \renewcommand\thefootnote{}\footnote{#1}%
  \addtocounter{footnote}{-1}%
  \endgroup
}

\usepackage[pagebackref=true,breaklinks=true,letterpaper=true,colorlinks,bookmarks=false]{hyperref}

\cvprfinalcopy 

\def\cvprPaperID{5619} 

\ifcvprfinal\fi

\title{
3D Packing for Self-Supervised Monocular Depth Estimation
}

\vspace{-2mm}
\author{
\large Vitor Guizilini
\quad
\large Rare\cb{s} Ambru\cb{s} 
\quad
\large Sudeep Pillai
\quad
\large Allan Raventos
\quad
\large Adrien Gaidon
\\
\vspace{-4mm}
\\
Toyota Research Institute (TRI)
\\
{\tt\small first.lastname@tri.global}
}

\begin{document}

\maketitle

\vspace*{-10mm}
\begin{abstract}
\vspace*{-2mm}
Although cameras are ubiquitous, robotic platforms typically rely on active sensors like LiDAR for direct 3D perception. In this work, we propose a novel self-supervised monocular depth estimation method combining geometry with a new deep network, \textit{PackNet}, learned only from unlabeled monocular videos. Our architecture leverages novel symmetrical packing and unpacking blocks to jointly learn to compress and decompress detail-preserving representations using 3D convolutions.
Although self-supervised, our method outperforms other self, semi, and fully supervised methods on the KITTI benchmark. The 3D inductive bias in PackNet enables it to scale with input resolution and number of parameters without overfitting, generalizing better on out-of-domain data such as the NuScenes dataset. Furthermore, it does not require large-scale supervised pretraining on ImageNet and can run in real-time. Finally, we release DDAD (Dense Depth for Automated Driving), a new urban driving dataset with more challenging and accurate depth evaluation, thanks to longer-range and denser ground-truth depth generated from high-density LiDARs mounted on a fleet of self-driving cars operating world-wide.$^\dagger$
\blfootnote{$^\dagger$Video: ~\href{https://www.youtube.com/watch?v=b62iDkLgGSI}{https://www.youtube.com/watch?v=b62iDkLgGSI}} 
\blfootnote{$^\dagger$Dataset: ~\href{https://github.com/TRI-ML/DDAD}{https://github.com/TRI-ML/DDAD}}
\blfootnote{$^\dagger$Code: ~\href{https://github.com/TRI-ML/packnet-sfm}{https://github.com/TRI-ML/packnet-sfm}}


\end{abstract}


\vspace{-5mm}
\section{Introduction}
\vspace{-1mm}

Accurate depth estimation is a key prerequisite in many robotics tasks, including perception, navigation, and planning. Depth from monocular camera configurations can provide useful cues for a wide array of tasks~\cite{kendall2018multi, lee2019spigan, manhardt2018roi, michels2005high}, producing dense depth maps that could complement or eventually replace expensive range sensors. However, learning monocular depth via direct
supervision requires ground-truth information from additional sensors and precise cross-calibration. Self-supervised methods do not suffer from these limitations, as they use geometrical constraints on image sequences as the sole source of supervision. In this work, we address the problem of jointly estimating scene structure and camera motion across RGB image sequences using a self-supervised deep network.

\begin{figure}[!t]
\vspace{-4mm}
    \centering

    \includegraphics[width=0.75\columnwidth,trim={0cm 0mm 35cm 1cm},clip, height=4.0cm]{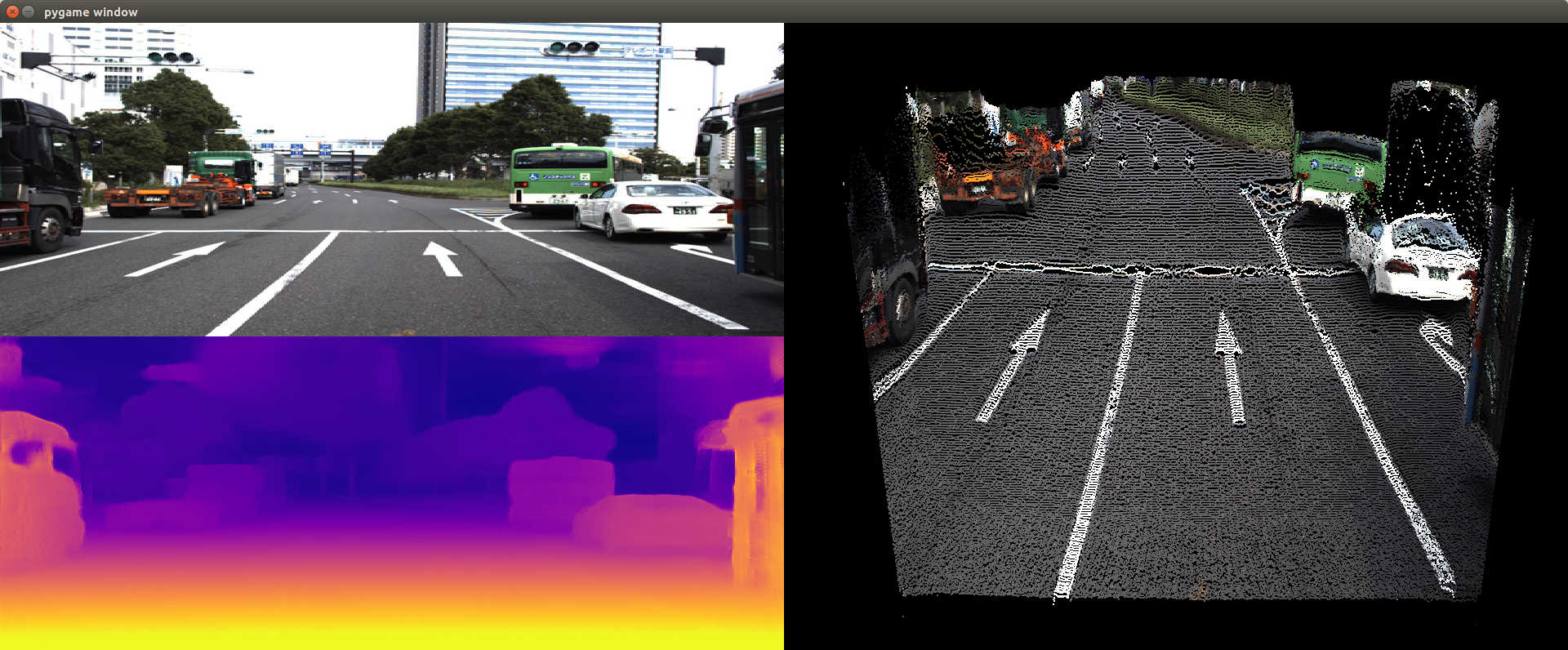} \\
    \includegraphics[width=0.75\columnwidth,trim={35cm 0mm 0mm 1cm},clip, height=3.1cm]{latex/figures/ddad/16-38-21.jpg}

    \caption{\textbf{Example metrically accurate PackNet prediction} (map and textured point cloud) on our DDAD dataset. 
    }
    
    \label{fig:intro-fig}
\vspace{-7mm}
\end{figure}

While recent works in self-supervised monocular depth estimation have mostly focused on engineering the loss function~\cite{casser2018depth,mahjourian2018unsupervised,yin2018geonet,zou2018dfnet}, we show that performance critically depends on the model architecture, in line with the observations of~\cite{kolesnikov2019revisiting} for other self-supervised tasks.
Going beyond image classification models like \textit{ResNet}~\cite{he2016deep}, our \textbf{main contribution} is a new convolutional network architecture, called \textit{PackNet}, for high-resolution self-supervised monocular depth estimation. We propose new packing and unpacking blocks that jointly
leverage 3D convolutions to learn representations that maximally propagate dense appearance and geometric information while still being able to run in real time.
Our \textbf{second contribution} is a novel loss that can optionally leverage the camera's velocity when available (e.g., from cars, robots, mobile phones) to solve the inherent scale ambiguity in monocular vision.
%
Our \textbf{third contribution} is a new dataset: \emph{Dense Depth for Automated Driving (DDAD)}. It leverages diverse logs from a fleet of well-calibrated self-driving cars equipped with cameras and high-accuracy long-range LiDARs. Compared to existing benchmarks, DDAD enables much more accurate depth evaluation at range, which is key for high resolution monocular depth estimation methods  (cf.~Figure~\ref{fig:intro-fig}).



Our experiments on the standard KITTI benchmark~\cite{geiger2013vision}, the recent NuScenes dataset \cite{nuscenes}, and our new proposed DDAD benchmark show that our self-supervised monocular approach \emph{i)} improves on the state of the art, especially at longer ranges; \emph{ii)} is competitive with fully supervised methods; \emph{iii)} generalizes better on unseen data; \emph{iv)} scales better with number of parameters, input resolution, and more unlabeled training data; \emph{v)} can run in real time at high resolution; and \emph{vi)} \textcolor{black}{does not require supervised pretraining on ImageNet to achieve state-of-the-art results; or test-time ground-truth scaling if velocity information is available at training time.}



\section{Related Work}
Depth estimation from a single image poses several challenges due to its ill-posed and ambiguous nature. However, modern convolutional networks have shown that it is possible to successfully leverage appearance-based patterns in large scale datasets in order to make accurate predictions.

\paragraph{Depth Network Architectures}
Eigen et al.~\cite{eigen2014depth} proposed one of the earliest works in convolutional-based depth estimation using a multi-scale deep network trained on {RGB-D} sensor data to regress the depth directly from single images. Subsequent works extended these network architectures to perform two-view stereo disparity estimation~\cite{mayer2016large} using techniques developed in the flow estimation literature~\cite{dosovitskiy2015flownet}. Following~\cite{dosovitskiy2015flownet, mayer2016large}, Umenhofer et al.~\cite{ummenhofer2017demon} applied these concepts to simultaneously train a depth and pose network to predict depth and camera ego-motion between successive unconstrained image pairs. 

Independently, dense pixel-prediction networks~\cite{bansal2017pixelnet,long2015fully,Yu_2017_CVPR} have made significant progress towards improving the flow of information between encoding and decoding layers. Fractional pooling ~\cite{fracpool} was introduced to amortize the rapid spatial reduction during downsampling. Lee et al.~\cite{lee2016generalizing} generalized the pooling function to allow the learning of more complex patterns, including linear combinations and learnable pooling operations. Shi et al.~\cite{shi2016real} used sub-pixel convolutions to perform Single-Image-Super-Resolution, synthesizing and super-resolving images beyond their input resolutions, while still operating at lower resolutions. Recent works~\cite{pillai2018superdepth,zhou2018unsupervised} in self-supervised monocular depth estimation use this concept to super-resolve estimates and further improve performance. 
Here, we go one step further and introduce new operations relying on 3D convolutions for learning to preserve and process spatial information in the features of encoding and decoding layers.

\paragraph{Self-Supervised Monocular Depth and Pose}
As supervised techniques for depth estimation advanced rapidly, the availability of target depth labels became challenging, especially for outdoor applications. To this end,~\cite{garg2016unsupervised,godard2017unsupervised} provided an alternative strategy involving training a monocular depth network with stereo cameras, without requiring ground-truth depth labels. By leveraging Spatial Transformer Networks~\cite{jaderberg2015spatial}, Godard et al~\cite{godard2017unsupervised} use stereo imagery to geometrically transform the right image plus a predicted depth of the left image into a synthesized left image. The loss between the resulting synthesized and original left images is then defined in a fully-differentiable manner, using a Structural Similarity~\cite{wang2004image} term and additional depth regularization terms, thus allowing the depth network to be self-supervised in an end-to-end fashion.

\vspace{2mm}
Following~\cite{godard2017unsupervised} and~\cite{ummenhofer2017demon}, Zhou et al.~\cite{zhou2017unsupervised} generalize this to self-supervised training in the \textit{purely} monocular setting, where a depth and pose network are simultaneously learned from unlabeled monocular videos. Several methods~\cite{casser2018depth,klodt2018supervising,mahjourian2018unsupervised,wang2018learning, yang2018deep,yin2018geonet,zhou2018unsupervised,zou2018dfnet} have advanced this line terms,of work by incorporatingthese methods, ad,ditional loss and constraints. All, however, take advantage of constraints in monocular Structure-from-Motion (SfM) training that only allow the estimation of depth and pose up to an unknown scale factor, and rely on the ground-truth LiDAR measu,rements to scale their depth estimates appropriately for evaluation purposes~\cite{zhou2017unsupervised}. Instead, in this work we show that, by simply using the instantaneous velocity of the camera during training, we are able to learn a \textit{scale-aware} depth and pose model, alleviating the impractical need to use LiDAR ground-truth depth measurements at test-time.

\section{Self-Supervised Scale-Aware SfM}
\label{sec:self-supervised-monocular-sfm}

In self-supervised monocular SfM training (Fig. \ref{fig:training-architecture}), we aim to learn: (i) a monocular depth model $f_D: I \to D$, that predicts the scale-ambiguous depth $\hat{D} = f_D(I(p))$ for every pixel $p$ in the target image $I$;
and (ii) a monocular ego-motion estimator $f_{\mathbf{x}}: (I_t,I_S) \to \mathbf{x}_{t \to S}$, that predicts the set of 6-DoF rigid transformations for all $s \in S$ given by $\mathbf{x}_{t \to s} = \begin{psmallmatrix}\mathbf{R} & \mathbf{t}\\ \mathbf{0} & \mathbf{1}\end{psmallmatrix} \in \text{SE(3)}$, between the target image $I_t$ and the set of source images $I_s \in I_S$ considered as part of the temporal context. In practice, we use the frames $I_{t-1}$ and $I_{t+1}$ as source images, although using a larger context is possible.
Note that in the case of monocular SfM both depth and pose are estimated up to an unknown scale factor, due to the inherent ambiguity of the photometric loss.

\subsection{Self-Supervised Objective}
\label{sec:preliminaries}
\begin{figure}
\vspace{-3mm}
    \centering
    \includegraphics[width=0.85\columnwidth,trim={0mm 10mm 0mm 0mm},clip]{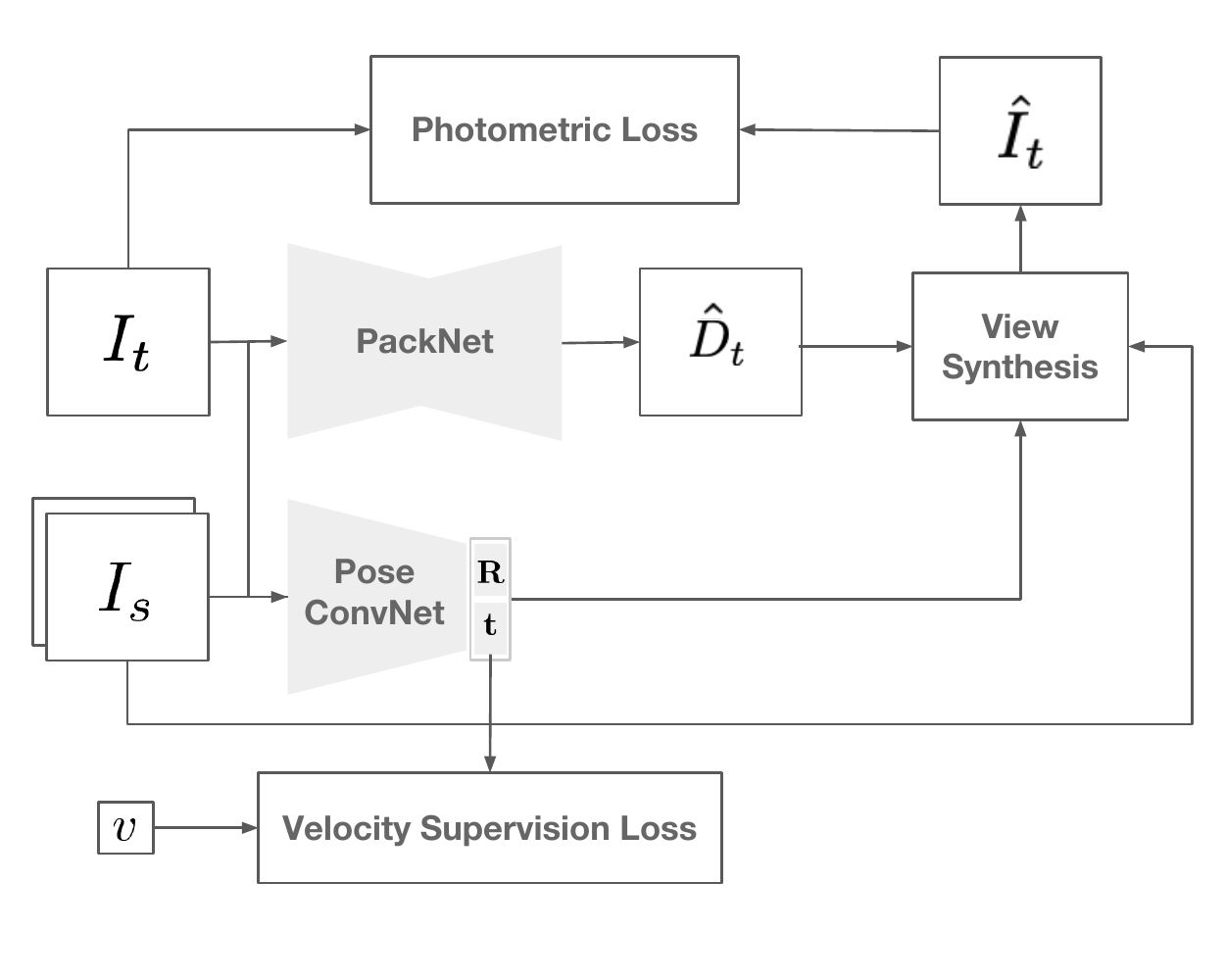}
    \label{fig:training_mono}
    \caption{\textbf{PackNet-SfM}: Our proposed scale-aware self-supervised monocular structure-from-motion architecture. We introduce \textit{PackNet} as a novel depth network, and optionally include weak velocity supervision at training time to produce \textit{scale-aware} depth and pose models.}
    \label{fig:training-architecture}
    \vspace{-3mm}
\end{figure}

Following the work of Zhou et al.~\cite{zhou2017unsupervised}, we train the depth and pose network simultaneously in a \textit{self-supervised} manner. In this work, however, we learn to recover the inverse-depth $f_d: I \to f^{-1}_D(I)$ instead, along with the ego-motion estimator $f_{\mathbf{x}}$. Similar to~\cite{zhou2017unsupervised}, the overall self-supervised objective consists of an appearance matching loss term $\mathcal{L}_p$ that is imposed between the synthesized target image $\hat{I}_t$ and the target image $I_t$, and a depth regularization term $\mathcal{L}_s$ that ensures edge-aware smoothing in the depth estimates $\hat{D}_t$. The objective takes the following form: \vspace{-1mm}
\begin{align}
    \mathcal{L}(I_t,\hat{I_t}) = \mathcal{L}_p(I_t,I_S) \odot \mathcal{M}_p \odot \mathcal{M}_t +  \lambda_1~\mathcal{L}_s(\hat{D}_t)
    \label{eq:overall-loss}
\end{align}
where $\mathcal{M}_t$ is a binary mask that avoids computing the photometric loss on the pixels that do not have a valid mapping, \textcolor{black}{and $\odot$ denotes element-wise multiplication}.
Additionally, $\lambda_1$ enforces a weighted depth regularization on the objective. The overall loss in Equation~\ref{eq:overall-loss} is averaged per-pixel, pyramid-scale and image batch during training. Fig.~\ref{fig:training-architecture} shows a high-level overview of our training pipeline. \vspace{2mm}

\textbf{Appearance Matching Loss.}~~Following~\cite{godard2017unsupervised,zhou2017unsupervised} the pixel-level similarity between the target image $I_t$ and the synthesized target image $\hat{I_t}$ is estimated using the Structural Similarity (SSIM)~\cite{wang2004image} term combined with an L1 pixel-wise loss term, inducing an overall photometric loss given by Equation~\ref{eq:loss-photo} below. \vspace{-1mm}
\begin{align}
  \resizebox{.9 \columnwidth}{!}{ 
    $\mathcal{L}_{p}(I_t,\hat{I_t}) = \alpha~\frac{1 - \text{SSIM}(I_t,\hat{I_t})}{2} + (1-\alpha)~\| I_t - \hat{I_t} \|$}
  \label{eq:loss-photo}
\end{align}
While multi-view projective geometry provides strong cues for self-supervision, errors due to 
\textcolor{black}{parallax 
in the scene} have an undesirable effect incurred on the photometric loss.  We mitigate these undesirable effects by calculating the minimum photometric loss per pixel for each source image in the context $I_S$, as shown in \cite{monodepth2}, so that:
\begin{equation}
\mathcal{L}_{p}(I_t, I_S) = \min_{I_S} \mathcal{L}_{p}(I_t,\hat{I}_t)
\end{equation}
The intuition is that the same pixel will not be occluded or out-of-bounds in all context images, and that the association with minimal photometric loss should be the correct one. Furthermore, we also mask out static pixels by removing those which have a \textit{warped} photometric loss $\mathcal{L}_{p}(I_t, \hat{I}_t)$ higher than their corresponding \textit{unwarped} photometric loss $\mathcal{L}_{p}(I_t, I_{s})$, calculated using the original source image without view synthesis. Introduced in \cite{monodepth2}, this auto-mask removes pixels whose appearance does not change between frames, which includes static scenes and dynamic objects with no relative motion, since these will have a smaller photometric loss when assuming no ego-motion.
\begin{equation}
\mathcal{M}_{p} = \min_{I_S} \mathcal{L}_{p}(I_t, I_s) > \min_{I_S} \mathcal{L}_{p}(I_t, \hat{I}_t)
\end{equation}

\textbf{Depth Smoothness Loss.}~~In order to regularize the depth in texture-less low-image gradient regions, we incorporate an edge-aware term (Equation~\ref{eq:loss-disp-smoothness}), similar to~\cite{godard2017unsupervised}. The loss is weighted for each of the pyramid-levels, and is decayed by a factor of 2 on down-sampling, starting with a weight of 1 for the $0^\text{th}$ pyramid level.
\begin{align}
  \mathcal{L}_{s}(\hat{D}_t) = | \delta_x \hat{D}_t | e^{-|\delta_x I_t|} + | \delta_y \hat{D}_t | e^{-|\delta_y I_t|}
  \label{eq:loss-disp-smoothness}
\end{align}

\subsection{Scale-Aware SfM}
\label{sec:velocity-scaling}
As previously mentioned, both the monocular depth and ego-motion estimators $f_d$ and $f_\mathbf{x}$ predict \textit{scale-ambiguous} values, due to the limitations of the monocular SfM training objective. In other words, the scene depth and the camera ego-motion can only be estimated up to an unknown and ambiguous scale factor. This is also reflected in the overall learning objective, where the photometric loss is agnostic to the \textit{metric} depth of the scene. Furthermore, we note that all previous approaches which operate in the self-supervised monocular regime~\cite{casser2018depth,garg2016unsupervised,godard2017unsupervised,mahjourian2018unsupervised} suffer from this limitation, and resort to artificially incorporating this scale factor at test-time, using LiDAR measurements. 
\vspace{2mm}

\textbf{Velocity Supervision Loss.}
Since instantaneous velocity measurements are ubiquitous in most mobile systems today, we show that they can be directly incorporated in our self-supervised objective to learn a metrically accurate and \textit{scale-aware} monocular depth estimator. During training, we impose an additional loss $\mathcal{L}_{v}$ between the magnitude of the pose-translation component of the pose network prediction $\hat{\mathbf{t}}$ and the measured instantaneous velocity scalar $v$ multiplied by the time difference between target and source frames $\Delta T_{t \to s}$, as shown below:
\vspace{-1mm}
\begin{align}
  \mathcal{L}_{v}({\hat{\mathbf{t}}_{t \to s}},v) = \Bigl|\|\hat{\mathbf{t}}_{t \to s} \| - |v| \Delta T_{t \to s}\Bigr|
  \label{eq:velocity-loss}
\end{align}
Our final scale-aware self-supervised objective loss $\mathcal{L}_{\text{scale}}$ from Equation~\ref{eq:overall-loss} becomes: 
\begin{align}
    \mathcal{L}_{\text{scale}}(I_t,\hat{I_t}, v) = \mathcal{L}(I_t,\hat{I_t}) +  \lambda_2~\mathcal{L}_{v}({\hat{\mathbf{t}}_{t \to s}},v)
    \label{eq:full-loss}
\end{align}
where $\lambda_2$ is a weight used to balance the different loss terms. This additional velocity loss allows the pose network to make metrically accurate predictions, subsequently resulting in the depth network also learning metrically accurate estimates to maintain consistency (cf. Section~\ref{sec:velocity-scaling-depth-experiments}).

\begin{figure}[!t]
    \centering
    \vspace{-5mm}
    {\subfloat[Packing]{
    \includegraphics[height=5.6cm]{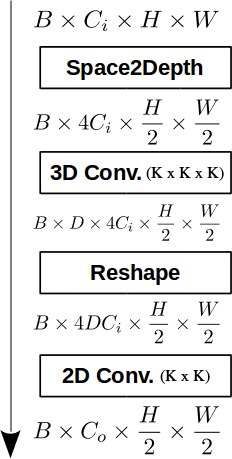}
    \label{fig:packing}}
    \qquad
    \subfloat[Unpacking]{
    \includegraphics[height=5.6cm]{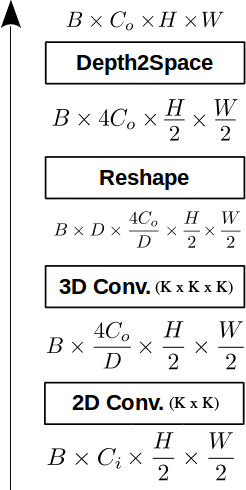}
    \label{fig:unpacking}}}
    \caption{\textbf{Proposed 3D packing and unpacking blocks.} \textit{Packing} replaces striding and pooling, while \textit{unpacking} is its symmetrical feature upsampling mechanism.}
    \label{fig:modules}
    \vspace{-3mm}
\end{figure}

\section{PackNet: 3D Packing for Depth Estimation}
\label{sec:packnet}

Standard convolutional architectures use aggressive striding and pooling to increase their receptive field size.
However, this potentially decreases model performance for tasks requiring fine-grained representations~\cite{fracpool,Zhang2018HartleySP}.
Similarly, traditional upsampling strategies \cite{ChenTho2017,DongSRDCNN} fail to propagate and preserve sufficient details at the decoder layers to recover accurate depth predictions.
In contrast, we propose a novel encoder-decoder architecture, called \textit{PackNet}, that introduces new 3D \textit{packing} and \textit{unpacking} blocks to \emph{learn} to \emph{jointly} preserve and recover important \emph{spatial} information for depth estimation.
This is in alignment with recent observations that information loss is not a necessary condition to learn representations capable of generalizing to different scenarios \cite{jacobsen2018irevnet}. In fact, progressive expansion and contraction in a fully invertible manner, without discarding ``uninformative" input variability, has been shown to increase performance in a wide variety of tasks \cite{behrmann2018invertible,dinh2016density,kingma2018glow}. 
We first describe the different blocks of our proposed architecture, and then proceed to show how they are integrated together in a single model for monocular depth estimation.




\subsection{Packing Block} 
\label{subsec:packing_block}

The \textit{packing} block (Fig.~\ref{fig:packing}) starts by folding the spatial dimensions of convolutional feature maps into extra feature channels via a \texttt{Space2Depth} operation~\cite{shi2016real}. The resulting tensor is at a reduced resolution, but in contrast to striding or pooling, this transformation is invertible and comes at no loss.
Next, we \textit{learn to compress} this concatenated feature space in order to reduce its dimensionality to a desired number of output channels.
As we show in our experiments (cf. Section~\ref{sec:ablation-studies}), 2D convolutions are not designed to directly leverage the tiled structure of this feature space. Instead, we propose to first \emph{learn to expand} this structured representation via a 3D convolutional layer. The resulting higher dimensional feature space is then flattened (by simple reshaping) before a final 2D convolutional contraction layer. This structured feature expansion-contraction, inspired by invertible networks~\cite{behrmann2018invertible,jacobsen2018irevnet} although we do not ensure invertibility, allows our architecture to dedicate more parameters to learn how to compress key spatial details that need to be preserved for high resolution depth decoding.



\begin{table}[!t]
\vspace{-4mm}
\renewcommand{\arraystretch}{1.05}
\centering
\resizebox{\linewidth}{!}{
\begin{tabular}{l|l|c|c}
\toprule
 & \textbf{Layer Description} & \textbf{K} & \textbf{Output Tensor Dim.} \\ \hline
\#0 & Input RGB image & & 3$\times$H$\times$W \\ 

\midrule
\multicolumn{4}{c}{\textbf{Encoding Layers}} \\ \hline
\#1 & Conv2d & 5 & 64$\times$H$\times$W \\
\#2 & Conv2d $\rightarrow$ Packing & 7 & 64$\times$H/2$\times$W/2 \\
\#3 & ResidualBlock (x2) $\rightarrow$ Packing & 3 & 64$\times$H/4$\times$W/4 \\
\#4 & ResidualBlock (x2) $\rightarrow$ Packing & 3 & 128$\times$H/8$\times$W/8 \\
\#5 & ResidualBlock (x3) $\rightarrow$ Packing & 3 & 256$\times$H/16$\times$W/16 \\
\#6 & ResidualBlock (x3) $\rightarrow$ Packing & 3 & 512$\times$H/32$\times$W/32 \\

\midrule
\multicolumn{4}{c}{\textbf{Decoding Layers}} \\ \hline
\#7 & Unpacking (\#6) $\rightarrow$ Conv2d ($\oplus$ \#5) & 3 &  512$\times$H/16$\times$W/16 \\

\#8 & Unpacking (\#7) $\rightarrow$ Conv2d ($\oplus$ \#4) & 3 & 256$\times$H/8$\times$W/8 \\
\textbf{\#9} & InvDepth (\#8) & 3 & 1$\times$H/8$\times$W/8 \\

\#10 & Unpacking (\#8) $\rightarrow$ Conv2d ($\oplus$ \#3 $\oplus$ Upsample(\#9)) & 3 & 128$\times$H/4$\times$W/4 \\
\textbf{\#11} & InvDepth {(\#10)} & 3 & 1$\times$H/4$\times$W/4 \\

\#12 & Unpacking (\#10) $\rightarrow$ Conv2d ($\oplus$ \#2 $\oplus$ Upsample(\#11)) & 3 & 64$\times$H/2$\times$W/2 \\
\textbf{\#13} & InvDepth {(\#12)} & 3 & 1$\times$H/2$\times$W/2 \\

\#14 & Unpacking (\#12) $\rightarrow$ Conv2d ($\oplus$ \#1 $\oplus$ Upsample(\#13)) & 3 & 64$\times$H$\times$W \\
\textbf{\#15} & InvDepth ({\#14}) & 3 &  1$\times$H$\times$W \\

\bottomrule
\end{tabular}}
\vspace{-2mm}
\caption{\textbf{Summary of our \textit{PackNet} architecture} for self-supervised monocular depth estimation. The \textit{Packing} and \textit{Unpacking} blocks are described in Fig.~\ref{fig:modules}, with kernel size $K=3$ and $D = 8$. \textit{Conv2d} blocks include  \textit{GroupNorm}~\cite{WuH18} with $G=16$ and ELU non-linearities~\cite{clevert2016fast}. \textit{InvDepth} blocks include a 2D convolutional layer with $K=3$ and sigmoid non-linearities. Each \textit{ResidualBlock} is a sequence of 3 2D convolutional layers with $K=3/3/1$ and ELU non-linearities, followed by \textit{GroupNorm} with $G=16$ and \textit{Dropout} \cite{dropout14} of 0.5 in the final layer. \textit{Upsample} is a nearest-neighbor resizing operation. Numbers in parentheses indicate input layers, with $\oplus$ as channel concatenation.
Bold numbers indicate the four inverse depth output scales.
}
\label{tab:packnet-arch}
\vspace{-3mm}
\end{table}

\subsection{Unpacking Block} 
\label{subsec:unpacking_block}

Symmetrically, the \textit{unpacking} block (Fig.~\ref{fig:unpacking}) \emph{learns to decompress and unfold} packed convolutional feature channels back into higher resolution spatial dimensions during the decoding process.
The unpacking block replaces convolutional feature upsampling, typically performed via nearest-neighbor or with learnable transposed convolutional weights. It is inspired by sub-pixel convolutions~\cite{shi2016real}, but adapted to reverse the 3D packing process that the features went through in the encoder.
First, we use a 2D convolutional layer to produce the required number of feature channels for a following 3D convolutional layer. Second, this 3D convolution learns to expand back the compressed spatial features. Third, these unpacked features are converted back to spatial details via a reshape and \texttt{Depth2Space} operation~\cite{shi2016real} to obtain a tensor with the desired number of output channels and target higher resolution.

\begin{figure}[t!]
    \vspace{-6mm}
    \captionsetup[subfloat]{justification=centering}
    \centering
    \small
    \subfloat[Input Image]{
    \includegraphics[width=0.32\columnwidth,trim={0cm 3.6cm 0cm 0cm},clip]{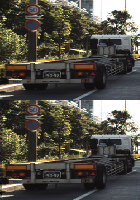}}
    \subfloat[Max Pooling + Bilinear Upsample]{
    \includegraphics[width=0.32\columnwidth,trim={0cm 0mm 0cm 3.6cm},clip]{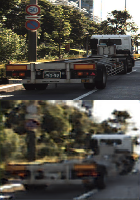}}
    \subfloat[Pack + Unpack]{
    \includegraphics[width=0.32\columnwidth,trim={0cm 0mm 0cm 3.6cm},clip]{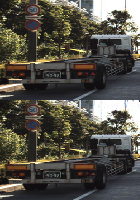}}
    \vspace{-2mm}
    \caption{\textbf{Image reconstruction} using different encoder-decoders: (b) standard max pooling and bilinear upsampling, each followed by 2D convolutions; (c) one packing-unpacking combination (cf.~Fig.~\ref{fig:modules}) with $D=2$. All kernel sizes are $K=3$ and $C=4$ for intermediate channels.  
}
    \label{fig:activations}
    \vspace{-3mm}
\end{figure}

\subsection{Detail-Preserving Properties}

\textcolor{black}{
In Fig. \ref{fig:activations}, we illustrate the detail-preserving properties of our packing / unpacking combination, showing we can get a near-lossless encoder-decoder for single image reconstruction by minimizing the L1 loss. We train a simple network composed of one packing layer followed by a symmetrical unpacking one and show it is able to almost exactly reconstruct the input image (final loss of $0.0079$), including sharp edges and finer details. In contrast, a comparable baseline replacing packing / unpacking with max pooling / bilinear upsampling (and keeping the 2D convolutions)
is only able to learn a blurry reconstruction (final loss of $0.063$). This highlights how \emph{PackNet} is able to learn more complex features by preserving spatial and appearance information end-to-end throughout the network.
}



\subsection{Model Architecture}
\label{sec:packnet-architecture}

Our \textit{PackNet} architecture for self-supervised monocular depth estimation is detailed in Table~\ref{tab:packnet-arch}.
Our symmetrical encoder-decoder architecture incorporates several packing and unpacking blocks, and is supplemented with skip connections~\cite{mayer2016large} to facilitate the flow of information and gradients throughout the network. 
%
%
%
The decoder produces intermediate inverse depth maps that are upsampled before being concatenated with their corresponding skip connections and unpacked feature maps. These intermediate inverse depth maps are also used at training time in the loss calculation, after being upsampled to to the full output resolution using nearest neighbors interpolation.




\section{Experiments}

\subsection{Datasets}
\label{sec:datasets-and-evaluation-metrics}

\textbf{KITTI~\cite{geiger2013vision}.}~The KITTI benchmark is the de facto standard for depth evaluation. More specifically, we adopt the training protocol used in Eigen et al.~\cite{eigen2014depth}, with Zhou et al.'s~\cite{zhou2017unsupervised} pre-processing to remove static frames. This results in 39810 images for training,  4424 for validation and 697 for evaluation. We also consider the improved ground-truth depth maps from \cite{gtkitti} for evaluation, which uses 5 consecutive frames to accumulate LiDAR points and stereo information to handle moving objects, resulting in 652 high-quality depth maps. 

\textbf{DDAD (Dense Depth for Automated Driving).} As one of our contributions, we release a diverse dataset of urban, highway, and residential scenes curated from a global fleet of self-driving cars. It contains 17,050 training and 4,150 evaluation frames with ground-truth depth maps generated from dense LiDAR measurements using the Luminar-H2 sensor. This new dataset is a more realistic and challenging benchmark for depth estimation, as it is diverse and captures precise structure across images ($30k$ points per frame) at longer ranges (up to $200m$ vs $80m$ for previous datasets). See supplementary material for more details.

\textbf{NuScenes~\cite{nuscenes}.}~To assess the generalization capability of our approach w.r.t.~previous ones, we evaluate KITTI models (without fine-tuning) on the official NuScenes validation dataset of 6019 front-facing images with ground-truth depth maps generated by LiDAR reprojection. 



\textbf{CityScapes~\cite{cordts2016cityscapes}.}~We also experiment with pretraining our monocular networks on the CityScapes dataset, before fine-tuning on the KITTI dataset. This also allows us to explore the scalability and generalization performance of different models, as they are trained with increasing amounts of unlabeled data. A total of $88250$ images were considered as the training split for the CityScapes dataset, \textcolor{black}{using the same training parameters as KITTI for $20$ epochs.}

\subsection{Implementation Details}
\label{sec:implementation-details}
We use PyTorch~\cite{paszke2017automatic} with all models trained across 8 Titan V100 GPUs. We use the Adam optimizer~\cite{kingma2014adam}, with $\beta_1=0.9$ and $\beta_2=0.999$. The monocular depth and pose networks are trained for $100$ epochs, with a batch size of 4 and initial depth and pose learning rates of $2 \cdot 10^{-4}$ and $5 \cdot 10^{-4}$ respectively. Training sequences are generated using a stride of 1, meaning that the previous $t-1$, current $t$, and posterior $t+1$ images are used in the loss calculation. As training proceeds, the learning rate is decayed every 40 epochs by a factor of 2. We set the SSIM weight to $\alpha=0.85$, the depth regularization weight to $\lambda_1=0.001$ and, where applicable, the velocity-scaling weight to $\lambda_2=0.05$. 

\textbf{Depth Network.}~~Unless noted otherwise, we use our \textit{PackNet} architecture as specified in Table~\ref{tab:packnet-arch}. During training, all four inverse depth output scales are used in the loss calculation, and at test-time only the final output scale is used, after being resized to the full ground-truth depth map resolution using nearest neighbor interpolation.

\textbf{Pose Network.}~~We use the architecture proposed by~\cite{zhou2017unsupervised} \textit{without} the explainability mask, which we found not to improve results. The pose network consists of 7 convolutional layers followed by a final $1 \times 1$ convolutional layer. The input to the network consists of the target view $I_t$ and the context views $I_S$, and the output is the set of 6 DOF transformations between $I_t$ and $I_s$, for $s \in S$. 


\subsection{Depth Estimation Performance}
\label{sec:depth-estimation-performance}

\begin{figure}[t!]
    \vspace{-3mm}
    \small
    \centering
    \includegraphics[trim={35cm 0mm 0mm 1cm},clip,width=4cm,height=2.1cm]{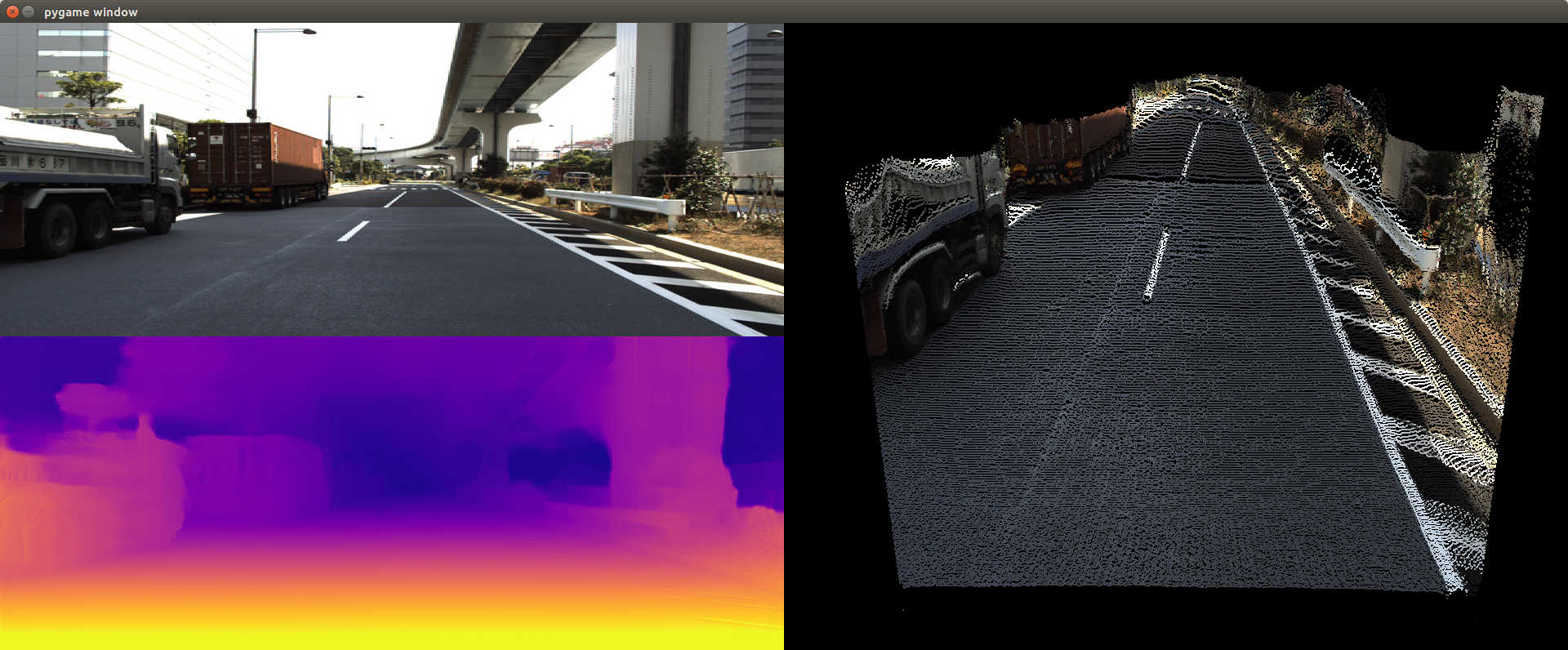}
    \hspace{-2mm}
    \includegraphics[trim={35cm 0mm 0mm 1cm},clip,width=4cm,height=2.1cm]{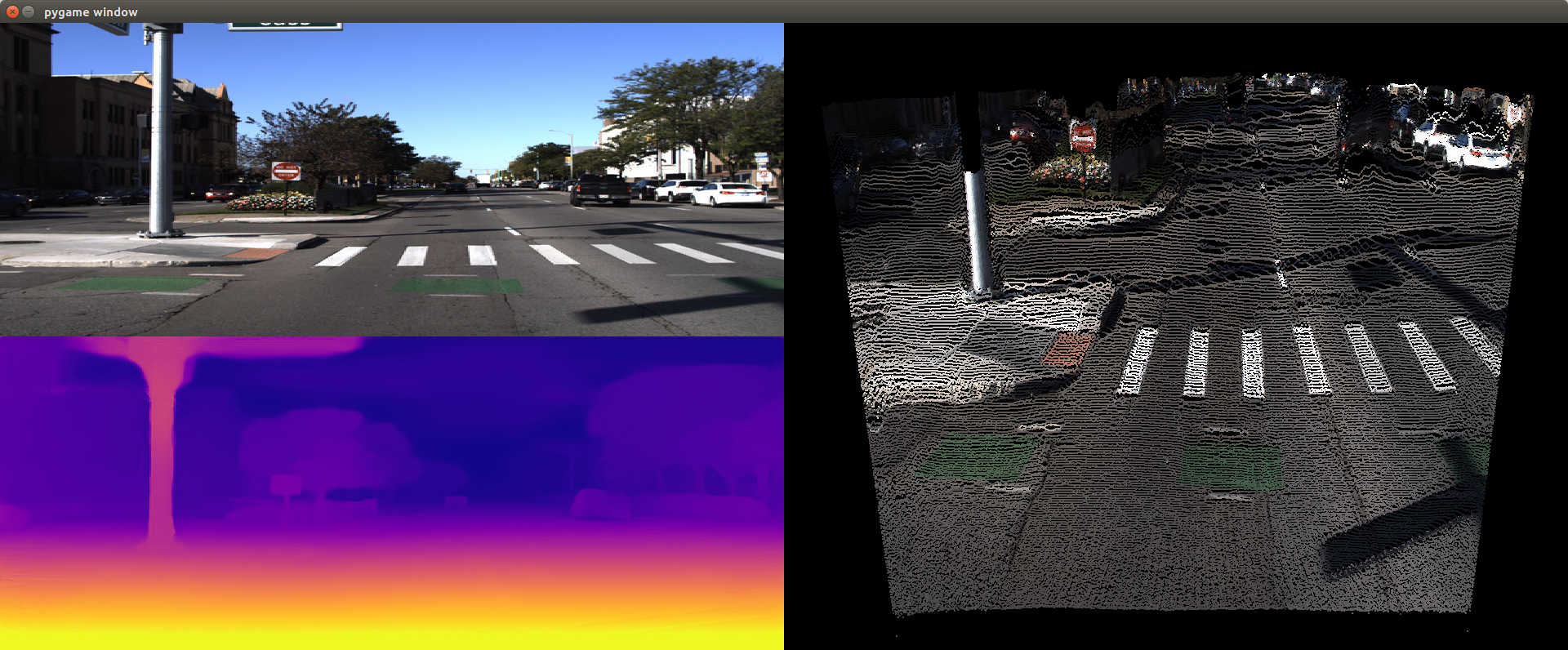}
    \\ \vspace{-3mm}
    \hspace{0.05mm}
    \includegraphics[trim={35cm 0mm 0mm 1cm},clip,width=4cm,height=2.1cm]{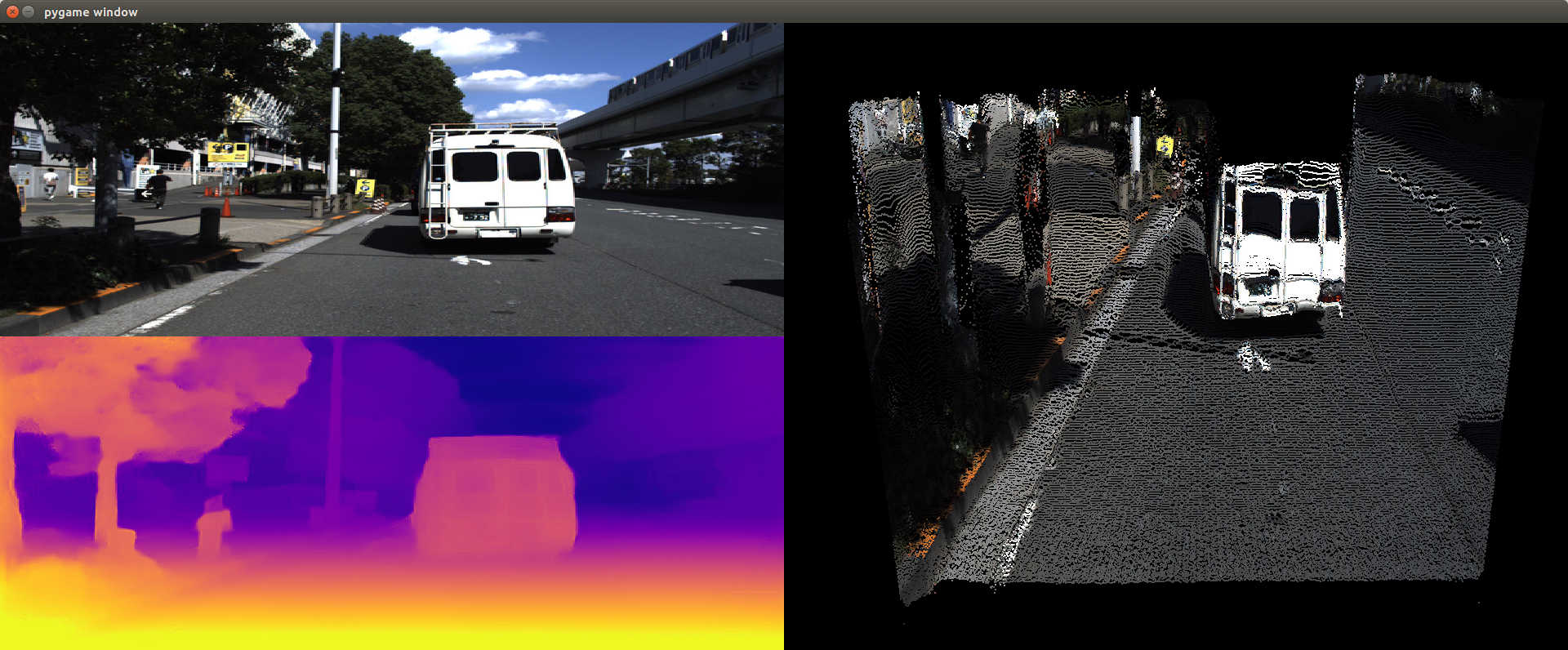}
    \hspace{-2mm}
    \includegraphics[trim={35cm 0mm 0mm 1cm},clip,width=4cm,height=2.1cm]{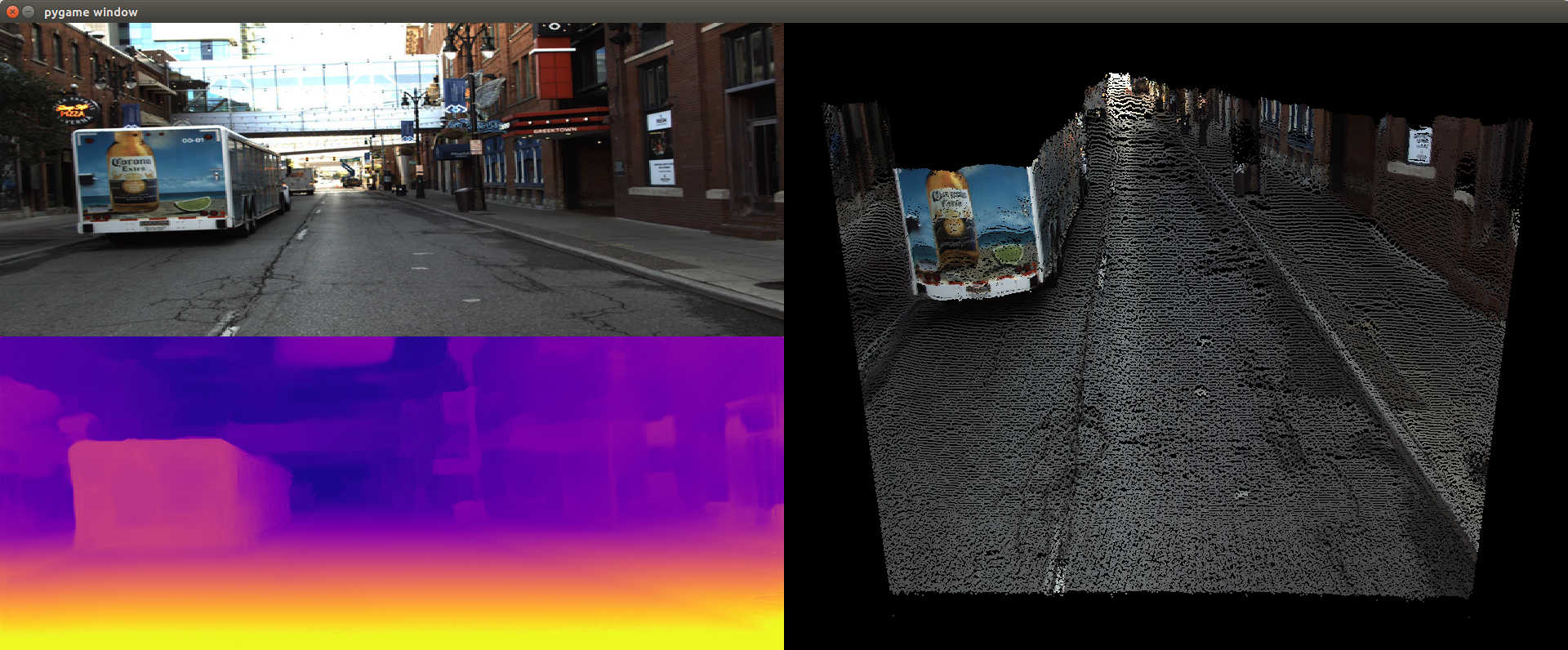}
    \vspace{-2mm}
    \caption{\textbf{PackNet pointcloud} reconstructions on DDAD.}
    \label{fig:ddad}
    \vspace{1mm}
\end{figure}

\begin{figure}[t!]
    \centering
    \small
    \includegraphics[width=0.85\columnwidth]{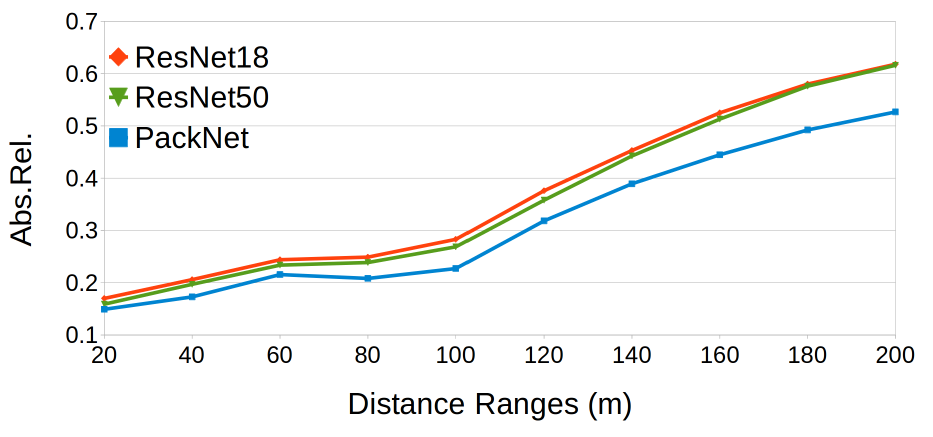}
\vspace{-2mm}
\caption{\textbf{Depth Evaluation on DDAD binned at different intervals}, calculated independently by only considering ground-truth depth pixels in that range (0-20m, 20-40m, ...). 
}
    \label{fig:depth_ranges}
\vspace{-3mm}
\end{figure}

First, we report the performance of our proposed monocular depth estimation method when considering longer distances, which is now possible due to the introduction of our new DDAD dataset. Depth estimation results using this dataset for training and evaluation, considering cumulative distances up to 200m, can be found in Fig.~\ref{fig:ddad} and Table \ref{tab:luminar}. Additionally, in Fig.~\ref{fig:depth_ranges} we present results for different depth intervals calculated independently. From these results we can see that our \emph{PackNet-SfM} approach significantly outperforms the state-of-the-art~\cite{monodepth2}, based on the \emph{ResNet} family, the performance gap consistently increasing when larger distances are considered.

Second, we evaluate depth predictions on KITTI using the metrics described in Eigen et al.~\cite{eigen2014depth}. We summarize our results in Table~\ref{table:depth-accuracy}, for the original depth maps from \cite{eigen2014depth} and the accumulated depth maps from \cite{gtkitti}, and illustrate their performance qualitatively in Fig.~\ref{fig:qualitative-depth}. In contrast to previous methods~\cite{casser2018depth,monodepth2} that predominantly focus on modifying the training objective, we show that our proposed \textit{PackNet} architecture can by itself bolster performance and establish a new state of the art for the task of monocular depth estimation, trained in the self-supervised monocular setting. 

Furthermore, we show that by simply introducing an additional source of unlabeled videos, such as the publicly available CityScapes dataset  (CS+K) \cite{cordts2016cityscapes}, we are able to further improve monocular depth estimation performance. As indicated by Pillai et al.~\cite{pillai2018superdepth}, we also observe an improvement in performance at higher image resolutions, which we attribute to the proposed network's ability to properly preserve and process spatial information \emph{end-to-end}. Our best results are achieved when injecting both more unlabeled data at training time and processing higher resolution input images, achieving performance comparable to semi-supervised \cite{kuznietsov2017semi} and fully supervised \cite{fu2018deep} methods.

\subsection{Scale-Aware Depth Estimation Performance}
\label{sec:velocity-scaling-depth-experiments}
Due to their inherent scale ambiguity, self-supervised monocular methods \cite{monodepth2,mahjourian2018unsupervised,zhou2017unsupervised} evaluate depth by scaling their estimates to the median ground-truth as measured via LiDAR. In Section~\ref{sec:velocity-scaling} we propose to also recover the metric scale of the scene from a single image by imposing a loss on the magnitude of the translation for the pose network output. Table \ref{table:depth-accuracy} shows that introducing this weak velocity supervision at training time allows the generation of \textit{scale-aware} depth models with similar performance as their unscaled counterparts, with the added benefit of not requiring ground-truth depth scaling (or even velocity information) at test-time. 
\textcolor{black}{
Another benefit of scale-awareness is that we can compose metrically accurate trajectories directly from the output of the pose network. Due to space constraints, we report pose estimation results in supplementary material.
}

\begin{table}[t!]
\vspace{-2mm}
\centering
{
\small
\setlength{\tabcolsep}{0.3em}
\renewcommand{\arraystretch}{0.9}
\begin{tabular}{lccccc}
\toprule
\textbf{Method} & 
Abs Rel &
Sq Rel &
RMSE &
RMSE$_{log}$ &
$\delta_{1.25}$ \\
\midrule
Monodepth2 (R18) & 0.381 & 8.387 & 21.277 & 0.371 & 0.587 \\
Monodepth2$^\ddagger$ (R18) & 0.213 & 4.975 & 18.051 & 0.340 & 0.761 \\
Monodepth2 (R50) & 0.324 & 7.348 & 20.538 & 0.344 & 0.615 \\
Monodepth2$^\ddagger$ (R50) & 0.198 & 4.504 & 16.641 & 0.318 & 0.781 \\
\textbf{PackNet-SfM} & \textbf{0.162} & \textbf{3.917} & \textbf{13.452} & \textbf{0.269} & \textbf{0.823} \\
\bottomrule
\end{tabular}
}
\vspace{-2mm}
\caption{\textbf{Depth Evaluation on DDAD}, for 640 x 384 resolution and distances up to 200m. While the \emph{ResNet} family heavily relies on large-scale supervised ImageNet \cite{Deng09imagenet} pretraining (denoted by $\ddagger$), \emph{PackNet} achieves significantly better results despite being trained from scratch.}

\vspace{-2mm}
\label{tab:luminar}
\end{table}


\begin{table*}[t!]
\renewcommand{\arraystretch}{0.87}
\centering
{
\small
\setlength{\tabcolsep}{0.3em}
\begin{tabular}{c|lcccccccccc}
\toprule
& \textbf{Method} &
Supervision & 
Resolution & 
Dataset &
Abs Rel &
Sq Rel &
RMSE &
RMSE$_{log}$ &
$\delta<1.25$ &
$\delta<1.25^2$ &
$\delta<1.25^3$\vspace{0.5mm}\\
\toprule

\parbox[t]{2mm}{\multirow{16}{*}{\rotatebox[origin=c]{90}{Original~\cite{eigen2014depth}}}}



& SfMLearner~\cite{zhou2017unsupervised} & M & 416 x 128 & CS + K & 0.198 & 1.836 & 6.565 & 0.275 & 0.718 & 0.901 & 0.960\\
& Vid2Depth~\cite{mahjourian2018unsupervised} & M & 416 x 128 & CS + K & 0.159 & 1.231 & 5.912 & 0.243 & 0.784 & 0.923 & 0.970 \\
& DF-Net~\cite{zou2018dfnet} & M & 576 x 160 & CS + K & 0.146 & 1.182 & 5.215 & 0.213 & 0.818 & 0.943 & 0.978 \\
& Struct2Depth~\cite{casser2018depth} & M & 416 x 128 & K & 0.141    & 1.026    & 5.291    &0.215    & 0.816    & 0.945    & 0.979\\
& \textcolor{black}{Zhou et al.}$^\ddagger$ ~\cite{zhou2019unsupervised} & M & 1248 x 384 & K & 0.121    & 0.837    & 4.945    &0.197    & 0.853    & 0.955    & 0.982\\

& Monodepth2$^\ddagger$~\cite{monodepth2} & M & 640 x 192 & K & 0.115 & 0.903 & 4.863 & 0.193 & 0.877 & 0.959 & 0.981\\
& Monodepth2$^\ddagger$~\cite{monodepth2} & M & 1024 x 320 & K & 0.115 & 0.882 & 4.701 & 0.190 & 0.879 & 0.961 & 0.982\\

\cmidrule{2-12} 

& \textbf{PackNet-SfM} & M & 640 x 192 & K & 0.111 & 0.785 & 4.601 & 0.189 & 0.878 & 0.960 & 0.982
\\
& \textbf{PackNet-SfM} & M+v & 640 x 192 & K & 0.111 & 0.829 & 4.788 & 0.199 & 0.864 & 0.954 & 0.980
\\
& \textbf{PackNet-SfM} & M & 640 x 192 & CS + K & 0.108 & \textbf{0.727} & {4.426} & 0.184 & 0.885 & 0.963 & \textbf{0.983}
\\
& \textbf{PackNet-SfM} & M+v & 640 x 192 & CS + K & 0.108 & 0.803 & 4.642 & 0.195 & 0.875 & 0.958 & 0.980
\\

\cmidrule{2-12} 

& \textbf{PackNet-SfM} & M & 1280 x 384 & K & 0.107 & 0.802 & 4.538 & 0.186 & 0.889 & 0.962 & 0.981
\\
& \textbf{PackNet-SfM} & M+v & 1280 x 384 & K & 0.107 & 0.803 & 4.566 & 0.197 & 0.876 & 0.957 & 0.979 
\\
& \textbf{PackNet-SfM} & M & 1280 x 384 & CS + K & {0.104} & {0.758} & \textbf{4.386} & \textbf{0.182} & \textbf{0.895} & \textbf{0.964} & {0.982} 
\\
& \textbf{PackNet-SfM} & M+v & 1280 x 384 & CS + K & \textbf{0.103} & 0.796 & 4.404 & 0.189 & 0.881 & 0.959 & 0.980
\\

\toprule
\toprule

\parbox[t]{2mm}{\multirow{11}{*}{\rotatebox[origin=c]{90}{Improved~\cite{gtkitti}}}}

& SfMLeaner~\cite{zhou2017unsupervised} & M & 416 x 128 & CS + K & 
0.176 & 1.532 & 6.129 & 0.244 & 0.758 & 0.921 & 0.971\\
& Vid2Depth~\cite{mahjourian2018unsupervised} & M & 416 x 128 & CS + K & 
0.134 & 0.983 & 5.501 & 0.203 & 0.827 & 0.944 & 0.981\\
& GeoNet~\cite{yin2018geonet} & M & 416 x 128 & CS + K & 
0.132 & 0.994 & 5.240 & 0.193 & 0.883 & 0.953 & 0.985\\
& DDVO~\cite{wang2018learning} & M & 416 x 128 & CS + K & 
0.126 & 0.866 & 4.932 & 0.185 & 0.851 & 0.958 & 0.986\\
& EPC++~\cite{epc++} & M & 640 x 192 & K & 
0.120 & 0.789 & 4.755 & 0.177 & 0.856 & 0.961 & 0.987\\
& Monodepth2$^\ddagger$~\cite{monodepth2} & M & 640 x 192 & K & 
0.090 & 0.545 & 3.942 & 0.137 & 0.914 & 0.983 & 0.995\\

\cmidrule{2-12} 

& Kuznietsov et al.$^\ddagger$ \cite{kuznietsov2017semi} & D & 621 x 187 & K & 
0.089 & 0.478 & 3.610 & 0.138 & 0.906 & 0.980 & 0.995\\

& DORN$^\ddagger$ \cite{fu2018deep} & D & 513 x 385 & K & 
{0.072} & \textbf{0.307} & \textbf{2.727} & 0.120 & 0.932 & 0.984 & 0.995\\

\cmidrule{2-12} 

& \textbf{PackNet-SfM} & M & 640 x 192 & K & 
0.078 & 0.420 & 3.485 & 0.121 & 0.931 & 0.986 & 0.996\\
& \textbf{PackNet-SfM} & M & 1280 x 384 & CS + K & 
\textbf{0.071} & 0.359 & 3.153 & \textbf{0.109} & \textbf{0.944} & \textbf{0.990} & \textbf{0.997}\\


& \textbf{PackNet-SfM} & M+v & 1280 x 384 & CS + K & 0.075 & 0.384 & 3.293 & 0.114 & 0.938 & 0.984 & 0.995 \\

\bottomrule

\end{tabular}
}
\caption{\textbf{Quantitative performance comparison of PackNet-SfM on the KITTI dataset} for distances up to 80m. For Abs Rel, Sq Rel, RMSE and RMSE$_{log}$ lower is better, and for $\delta < 1.25$, $\delta < 1.25^2$ and $\delta < 1.25^3$ higher is better. In the \emph{Dataset} column,  CS$+$K refers to pretraining on CityScapes (CS) and fine-tuning on KITTI (K). \text{M} refers to methods that train using monocular (M) images, and \text{M+v} refers to added velocity weak supervision (v), as shown in Section \ref{sec:velocity-scaling}. $^\ddagger$ indicates ImageNet~\cite{Deng09imagenet} pretraining. \textit{Original} uses raw depth maps from \cite{eigen2014depth} for evaluation, and \textit{Improved} uses annotated depth maps from \cite{gtkitti}. At test-time, all monocular methods (M) scale estimated depths with median ground-truth LiDAR information. Velocity-scaled (M+v) and supervised (D) methods are \textit{not} scaled in such way, since they are already metrically accurate.}
\label{table:depth-accuracy}
\end{table*}

\begin{figure*}[!b]
  \vspace{-10mm}
  \centering
  {
   {\renewcommand{\arraystretch}{0.2}
    \setlength{\tabcolsep}{0.2mm}
    \begin{tabular}{ccccc}
    Input image 
    & PackNet-SfM
    & Monodepth2 \cite{monodepth2}
    & DORN \cite{fu2018deep} 
    & SfMLearner \cite{zhou2017unsupervised} 
    \\
    \includegraphics[width=0.4\columnwidth,height=1.10cm]{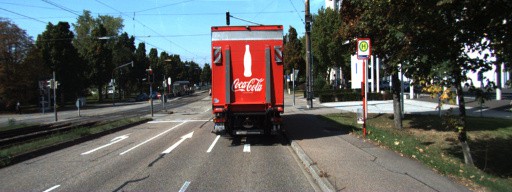}&
    \includegraphics[width=0.4\columnwidth,height=1.10cm]{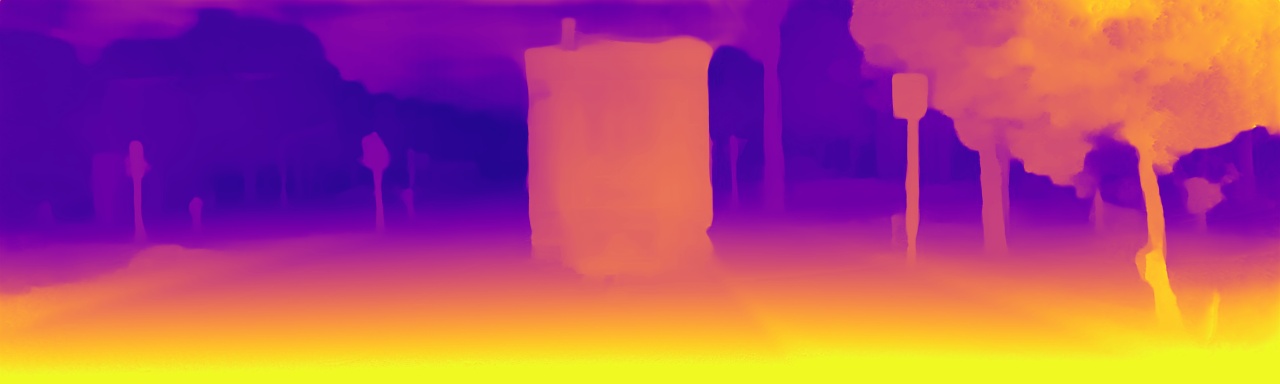}  &
    \includegraphics[width=0.4\columnwidth,height=1.10cm]{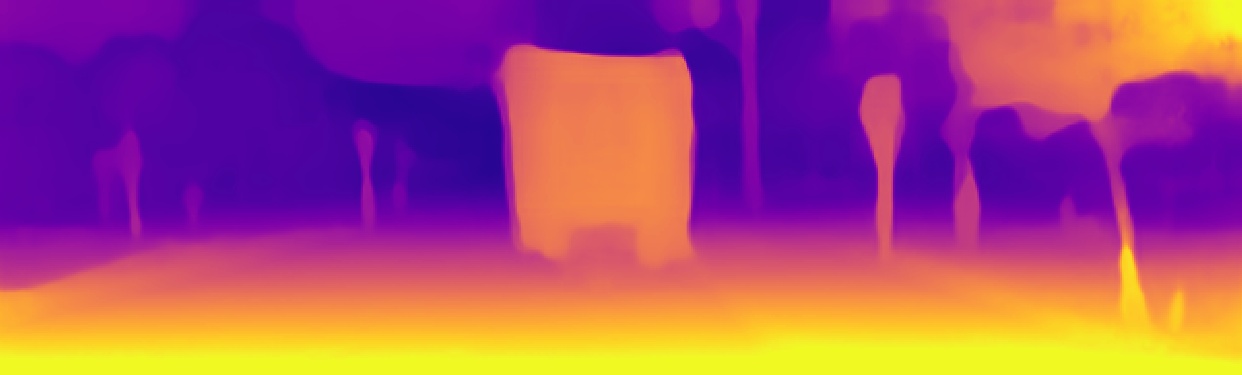}&
    \includegraphics[width=0.4\columnwidth,height=1.10cm]{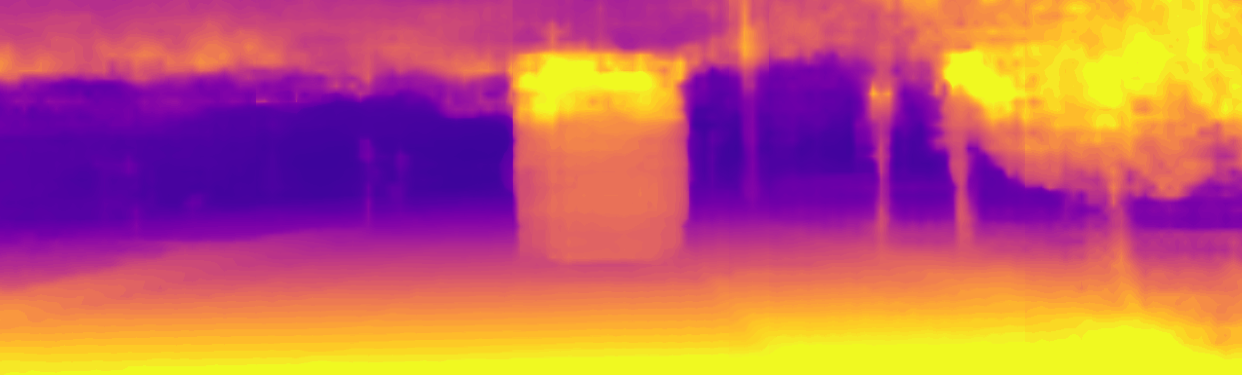}&
    \includegraphics[width=0.4\columnwidth,height=1.10cm]{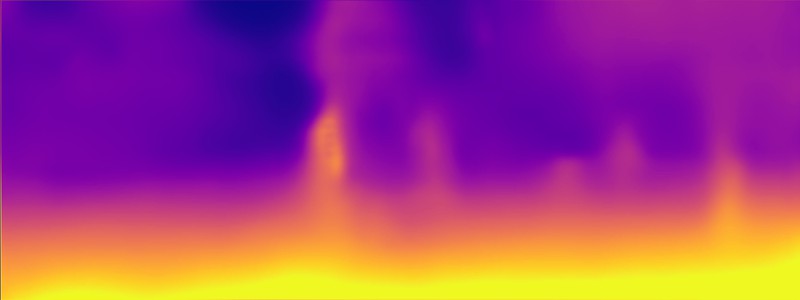}

    \\
    


    \includegraphics[width=0.4\columnwidth,height=1.10cm]{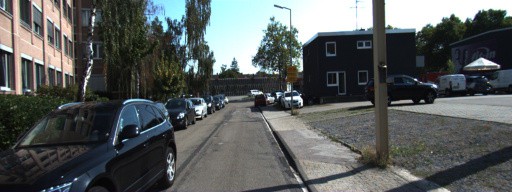}&
    \includegraphics[width=0.4\columnwidth,height=1.10cm]{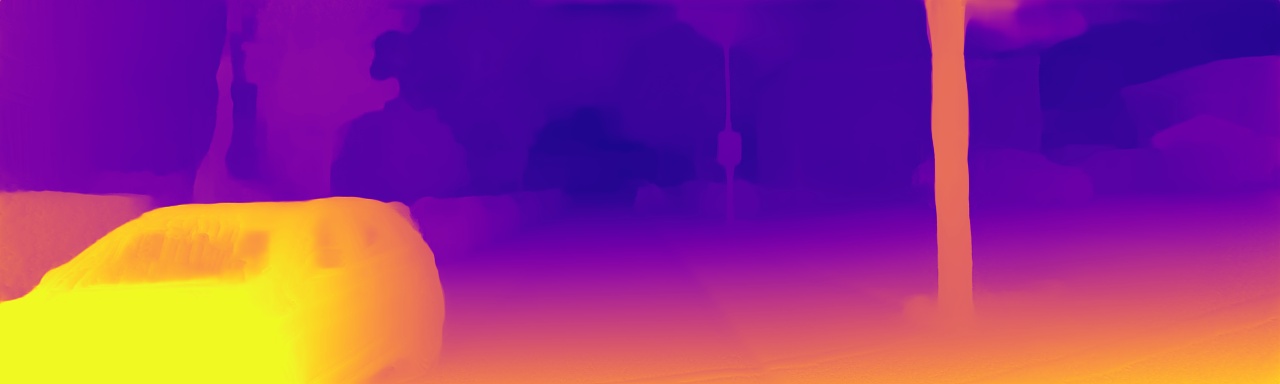}&
    \includegraphics[width=0.4\columnwidth,height=1.10cm]{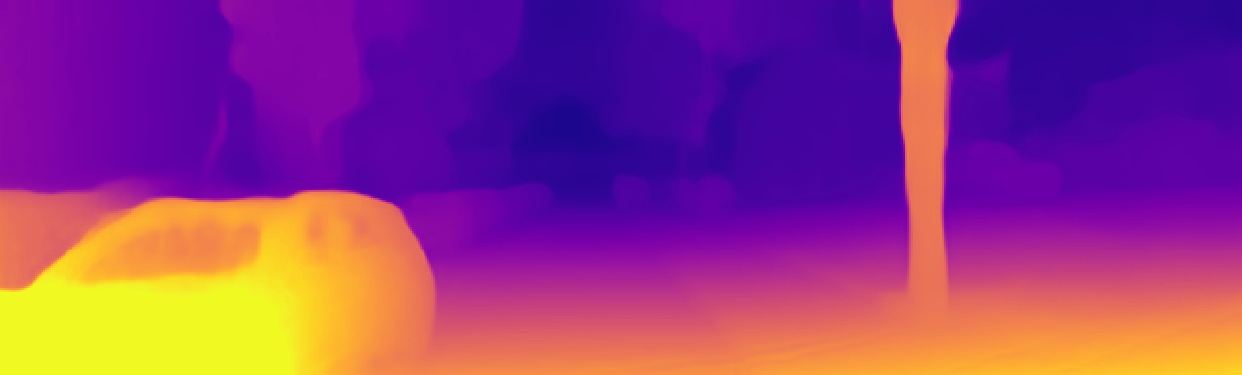}&
    \includegraphics[width=0.4\columnwidth,height=1.10cm]{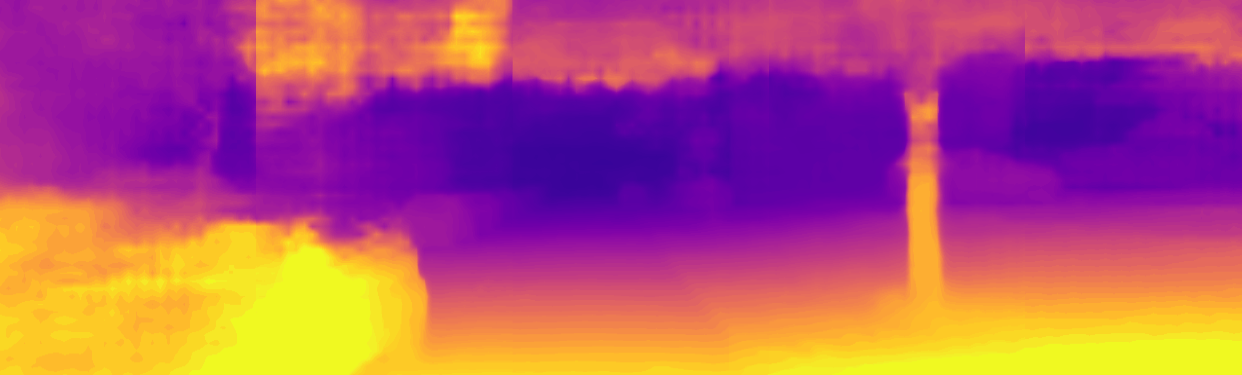}&
    \includegraphics[width=0.4\columnwidth,height=1.10cm]{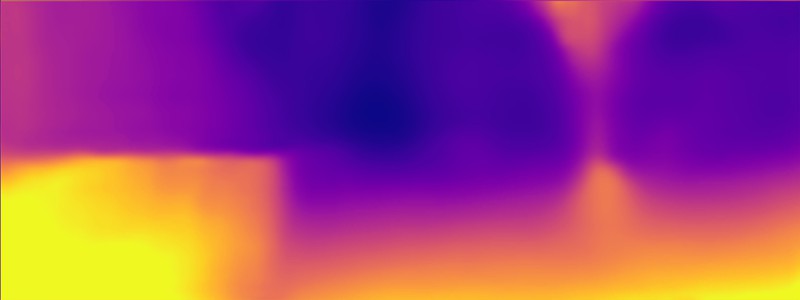}

    \\

    \includegraphics[width=0.4\columnwidth,height=1.10cm]{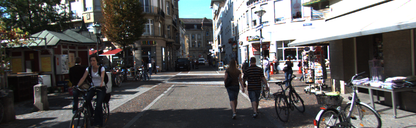}&
    \includegraphics[width=0.4\columnwidth,height=1.10cm]{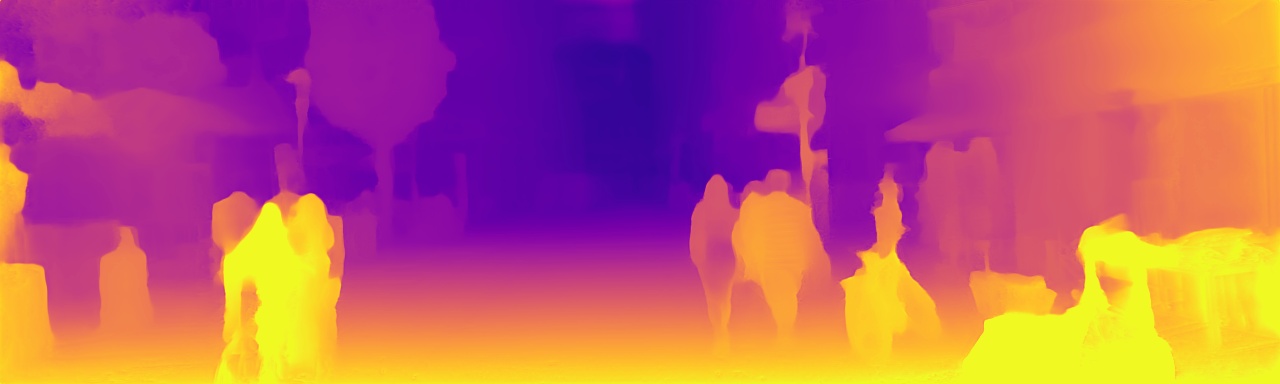}&
    \includegraphics[width=0.4\columnwidth,height=1.10cm]{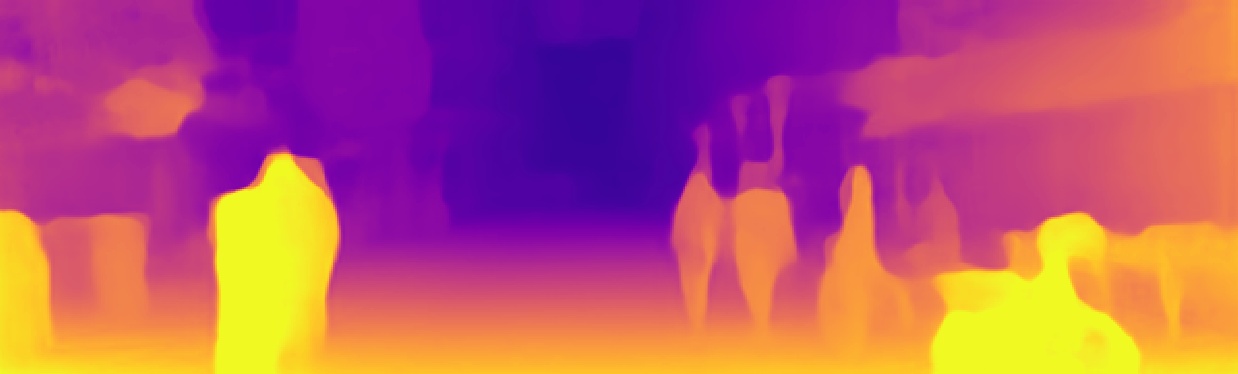} &
    \includegraphics[width=0.4\columnwidth,height=1.10cm]{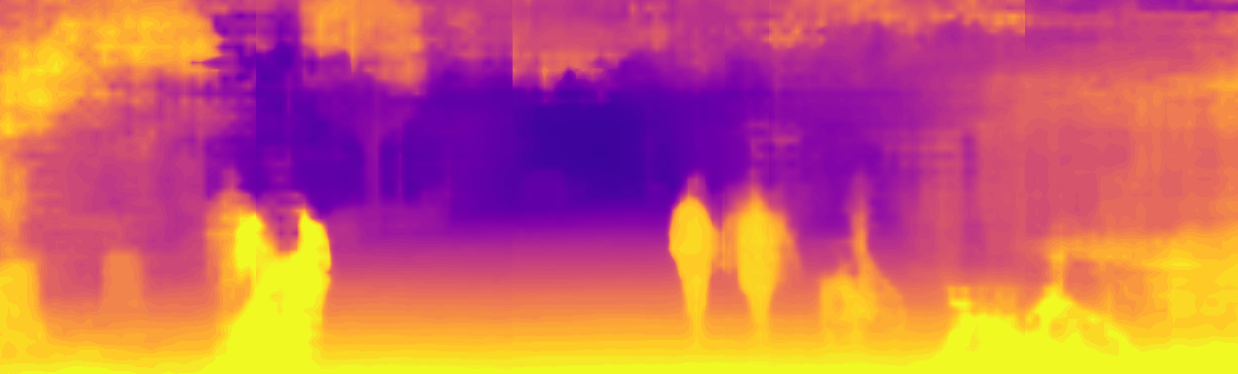}&
    \includegraphics[width=0.4\columnwidth,height=1.10cm]{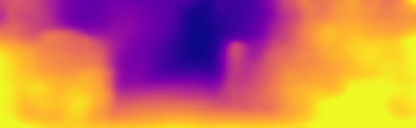}

    \\

\includegraphics[width=0.4\columnwidth,height=1.10cm]{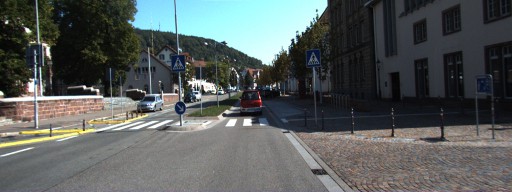}&
    \includegraphics[width=0.4\columnwidth,height=1.10cm]{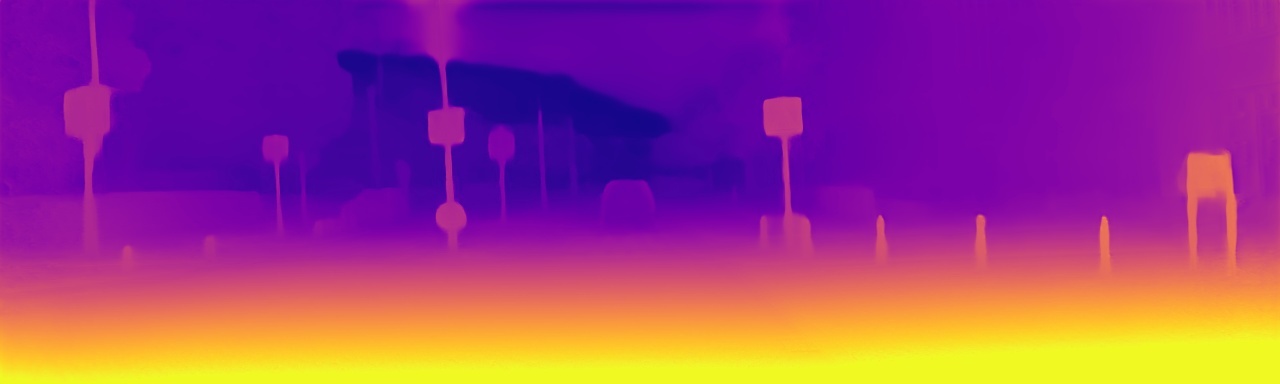}&
    \includegraphics[width=0.4\columnwidth,height=1.10cm]{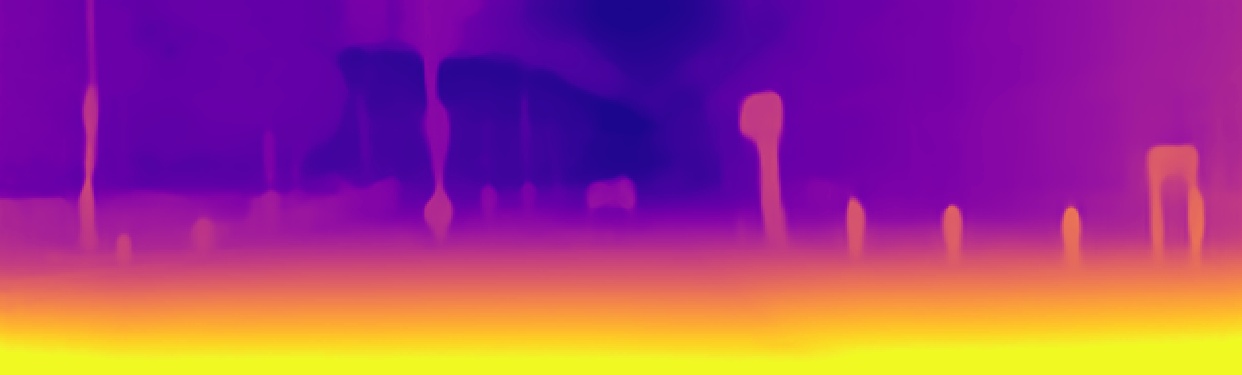}&
    \includegraphics[width=0.4\columnwidth,height=1.10cm]{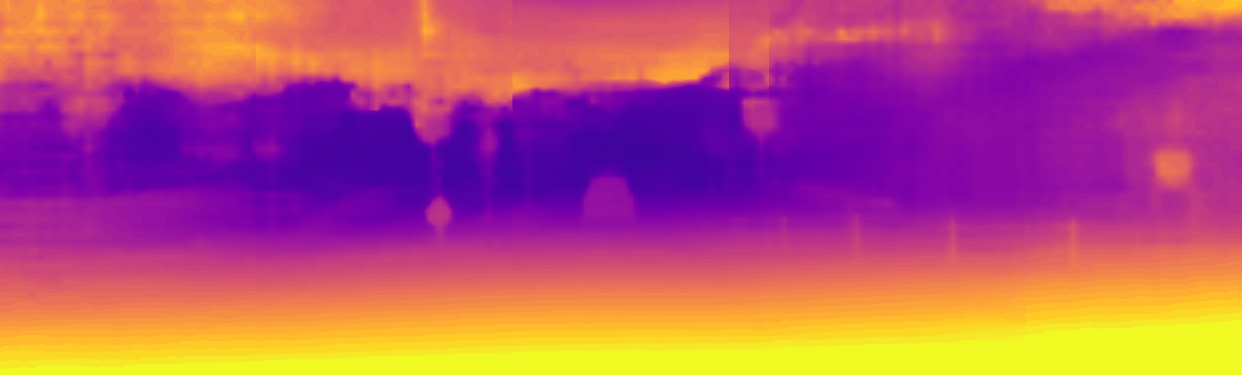}&
    \includegraphics[width=0.4\columnwidth,height=1.10cm]{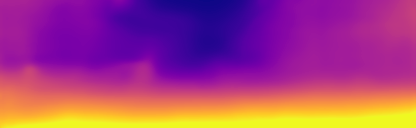}

    \\

    \includegraphics[width=0.4\columnwidth,height=1.10cm]{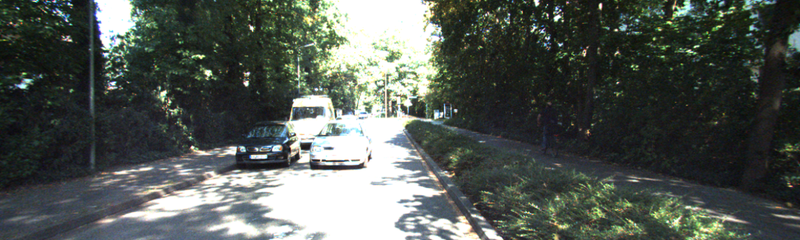}&
    \includegraphics[width=0.4\columnwidth,height=1.10cm]{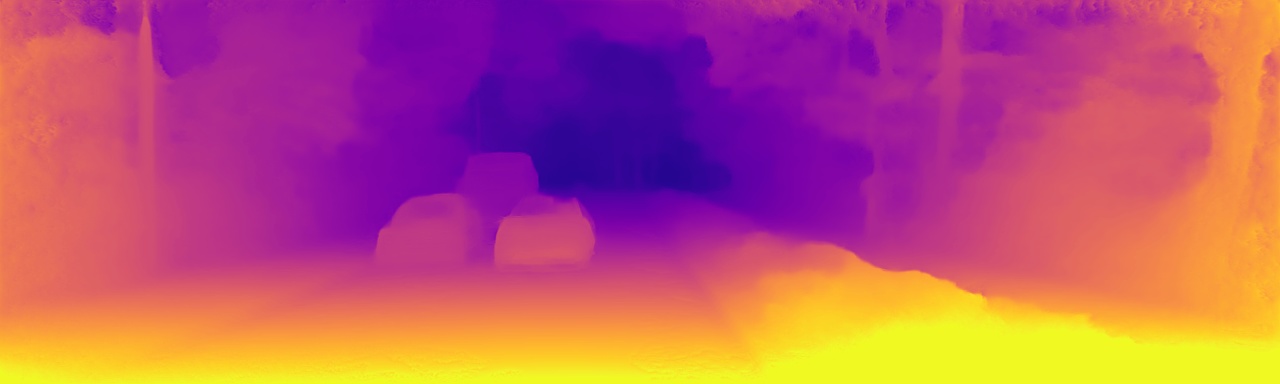}&
    \includegraphics[width=0.4\columnwidth,height=1.10cm]{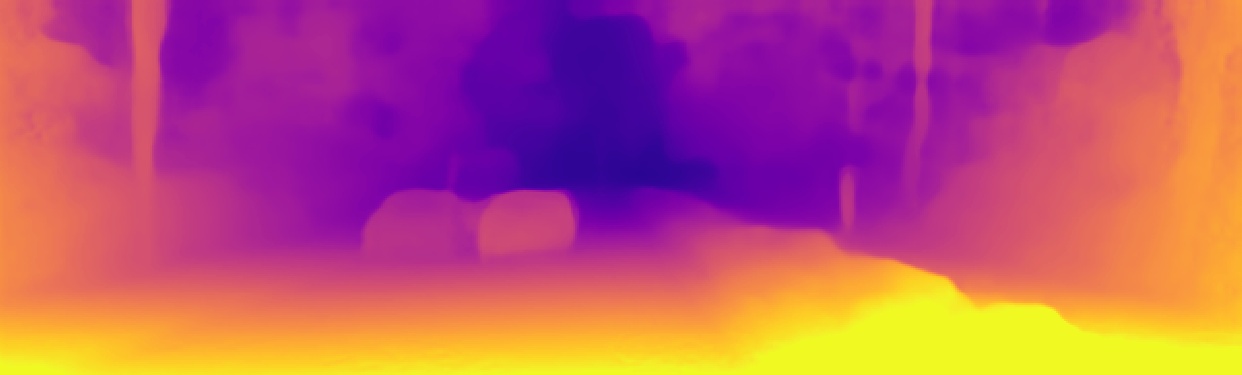}&
    \includegraphics[width=0.4\columnwidth,height=1.10cm]{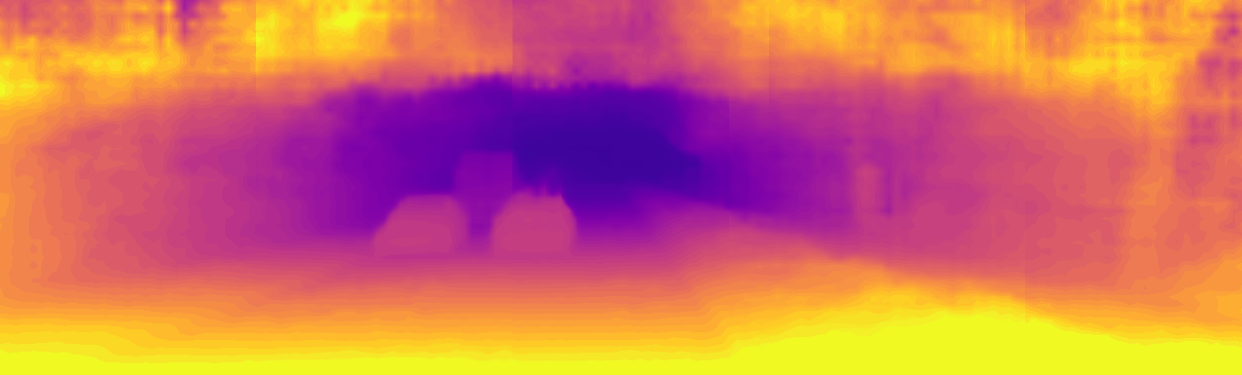}&    
    \includegraphics[width=0.4\columnwidth,height=1.10cm]{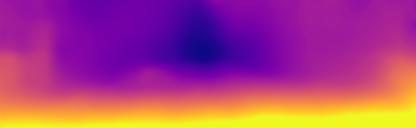}

    \\

    \includegraphics[width=0.4\columnwidth,height=1.10cm]{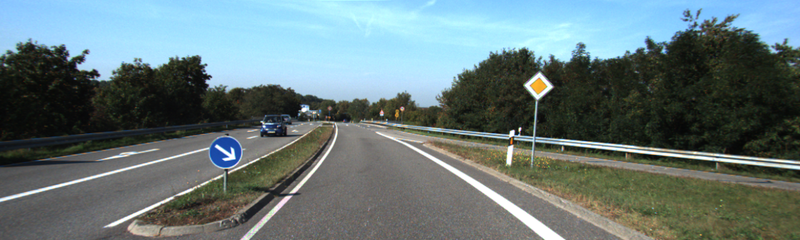}&
    \includegraphics[width=0.4\columnwidth,height=1.10cm]{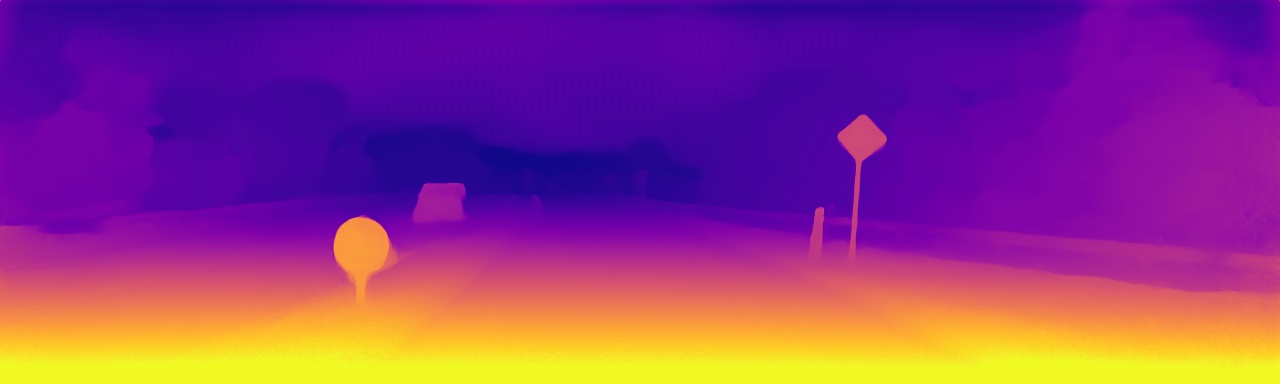}&
    \includegraphics[width=0.4\columnwidth,height=1.10cm]{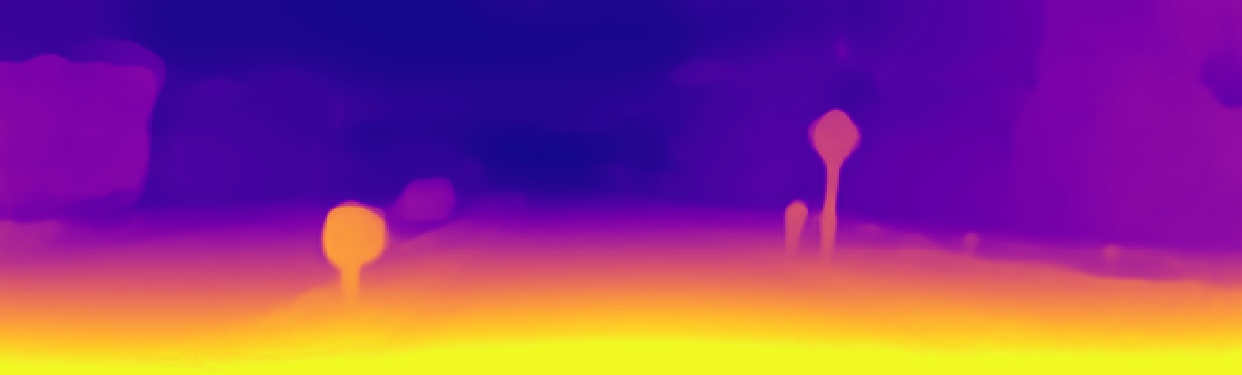}&
    \includegraphics[width=0.4\columnwidth,height=1.10cm]{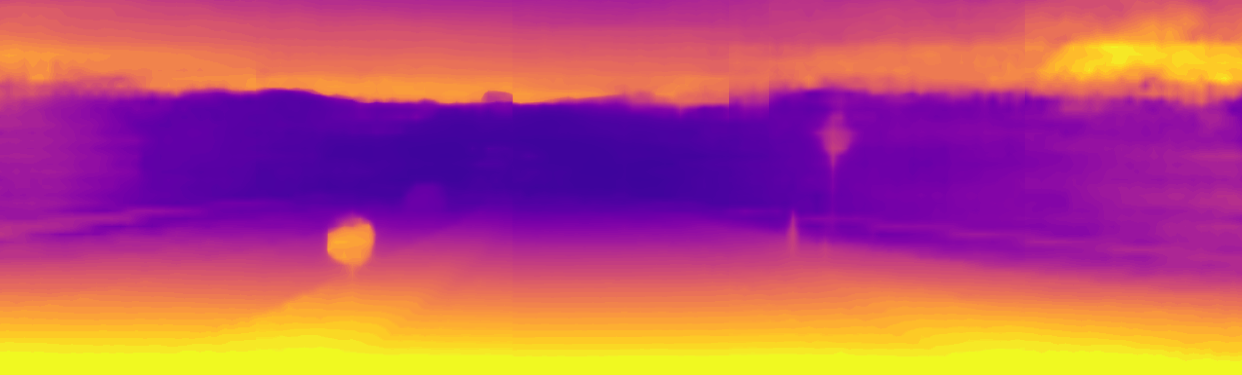}&
    \includegraphics[width=0.4\columnwidth,height=1.10cm]{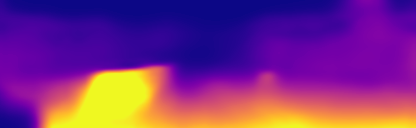}

    \\

\end{tabular}}}
  \caption{\textbf{Qualitative monocular depth estimation performance} comparing \textit{PackNet} with previous methods, on frames from the KITTI dataset (Eigen test split). Our method is able to capture sharper details and structure (e.g., on vehicles, pedestrians, and thin poles) thanks to the learned preservation of spatial information.
  }
  \label{fig:qualitative-depth}
\end{figure*}

\subsection{Network Complexity}

\begin{figure}[t!]
    \vspace{-3mm}
    \centering
    \small
    \includegraphics[width=0.91\columnwidth]{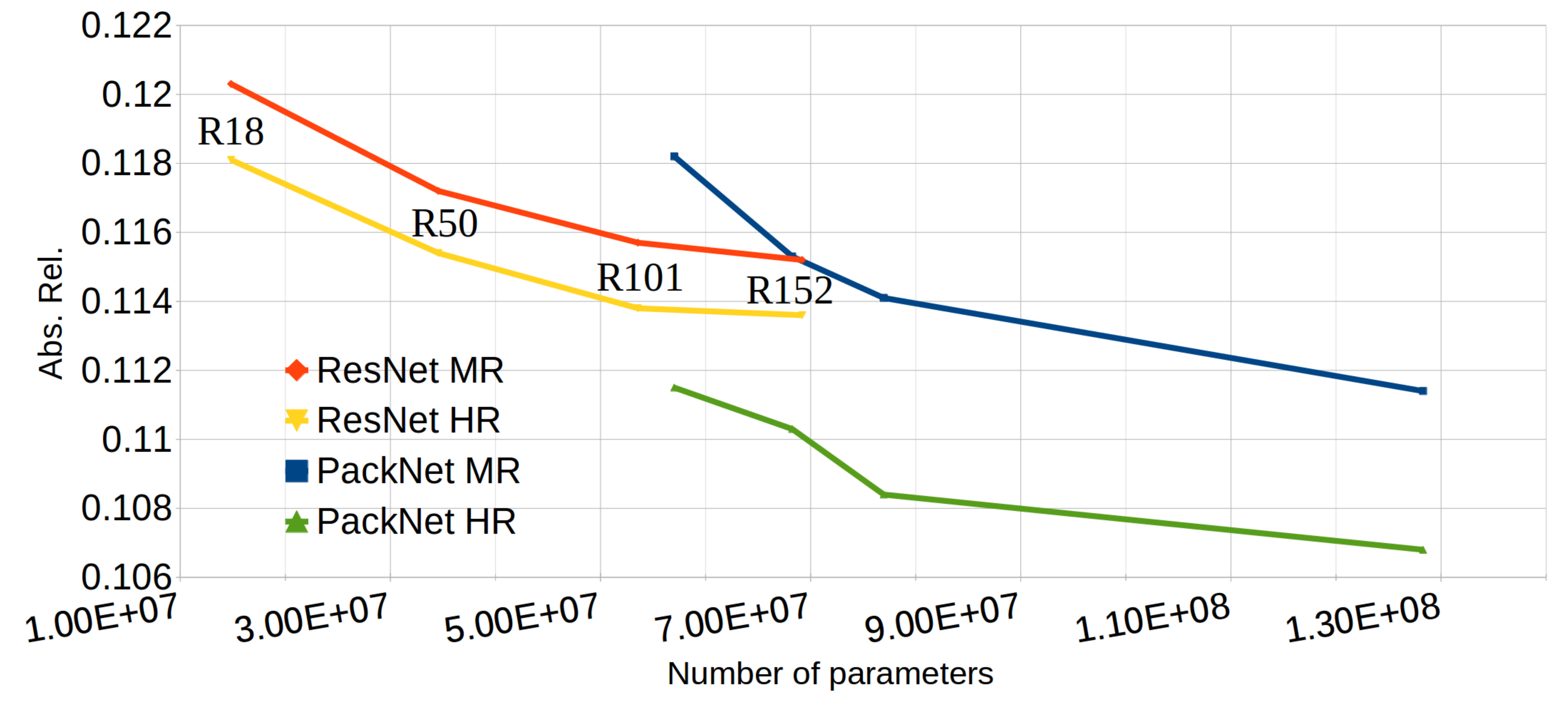}
    \caption{\textbf{Performance of different depth network architectures for varying numbers of parameters} on the original KITTI Eigen split \cite{eigen2014depth} with resolutions of 640 x 192 (MR) and 1280 x 384 (HR). While the \textit{ResNet} family plateaus at 70M parameters, the \textit{PackNet} family matches its performance at the same number of parameters for MR, outperforms it clearly for HR, and improves significantly with more parameters in both cases without overfitting.}
    \label{fig:parameters}    
    \vspace{-6mm}
\end{figure}

The introduction of packing and unpacking as alternatives to standard downsampling and upsampling operations increases the complexity of the network, due to the number of added parameters. To ensure that the gain in performance shown in our experiments is not only due to an increase in model capacity, we compare different variations of our \textit{PackNet} architecture (obtained by modifying the number of layers and feature channels) against available \textit{ResNet} architectures.
These results are depicted in Fig. \ref{fig:parameters} and show that, while the \textit{ResNet} family stabilizes with diminishing returns as the number of parameters increase, the \textit{PackNet} family matches its performance at around 70M parameters and further improves as more complexity is added. Finally, the proposed architecture (Table \ref{tab:packnet-arch}) reaches around 128M parameters with an inference time of 60ms on a Titan V100 GPU, \textcolor{black}{which can be further improved to $<$ 30ms using TensorRT \cite{tensorrt}}, making it suitable for real-time applications.

The \textit{PackNet} family is also consistently better at higher resolution, as it properly preserves and propagates spatial information between layers. In contrast, as reported in prior works~\cite{monodepth2}, \textit{ResNet} architectures do not scale well, with only minor improvements at higher resolution. 


\subsection{Ablation Studies}
\label{sec:ablation-studies}
To further study the performance improvements that \textit{PackNet} provides, we perform an ablative analysis on the different architectural components introduced, as depicted in Table \ref{table:packnet-ablation}.
We show that the base architecture, without the proposed packing and unpacking blocks, already produces a strong baseline for the monocular depth estimation task. 
\textcolor{black}{
The introduction of packing and unpacking boosts depth estimation performance, especially as more 3D convolutional filters are added, with new state-of-the-art results being achieved by the architecture described in Table \ref{tab:packnet-arch}.
}



As mentioned in~\cite{fu2018deep,monodepth2}, \textit{ResNet} architectures highly benefit from ImageNet pretraining, since they were originally developed for classification tasks. Interestingly, we also noticed that the performance of pretrained \textit{ResNet} architectures degrades in longer training periods, due to catastrophic forgetting that leads to overfitting. The proposed \textit{PackNet} architecture, on the other hand, achieves state-of-the-art results from randomly initialized weights, and can be further improved by self-supervised pretraining on other datasets, thus properly leveraging the large-scale availability of unlabeled information thanks to its structure.

\begin{table}[t!]
\vspace{-4mm}
\centering
{
\small
\renewcommand{\arraystretch}{0.85}
\setlength{\tabcolsep}{0.2em}
\begin{tabular}{lccccc}
\toprule
\textbf{Depth Network} & 
Abs Rel &
Sq Rel &
RMSE &
RMSE$_{log}$ &
$\delta_{1.25}$ \vspace{0.5mm}\\
\midrule
ResNet18 & 0.133 & 1.023 & 5.123 & 0.211 & 0.845 
\\
ResNet18$^\ddagger$ & 0.120 & 0.896 & 4.869 & 0.198 & 0.868 
\\
ResNet50 & 0.127 & 0.977 & 5.023 & 0.205 & 0.856 
\\
ResNet50$^\ddagger$ & 0.117 & 0.900 & 4.826 & 0.196 & 0.873 
\\
\midrule
PackNet & \multirow{2}{*}{0.122} & \multirow{2}{*}{0.880} & \multirow{2}{*}{4.816} & \multirow{2}{*}{0.198} & \multirow{2}{*}{0.864} \\
\small{(w/o pack/unpack)} & 
\\
PackNet ($D=0$) & 0.121 & 0.922 & 4.831 & 0.195 & 0.869  \\
PackNet ($D=2$) & 0.118 & 0.802 & 4.656 & 0.194 & 0.868  \\
PackNet ($D=4$) & 0.113 & 0.818 & 4.621 & 0.190 & 0.875  \\
PackNet ($D=8$) & 0.111 & 0.785 & 4.601 & 0.189 & 0.878  \\

\bottomrule
\end{tabular}
}
\vspace{-1mm}
\caption{
\textcolor{black}{
\textbf{Ablation study on the PackNet architecture}, on the standand KITTI benchmark for 640 x 192 resolution. \textit{ResNetXX} indicates that specific architecture~\cite{he2016deep} as encoder, with and without ImageNet~\cite{Deng09imagenet} pretraining (denoted with $\ddagger$). We also show results with the proposed \textit{PackNet} architecture, first without packing and unpacking (replaced respectively with convolutional striding and bilinear upsampling) and then with increasing numbers of 3D convolutional filters ($D=0$ indicates no 3D convolutions \textcolor{black}{and the corresponding reshape operations}).
}
}
\label{table:packnet-ablation}
\end{table}

\begin{table}[t!]
\vspace{-1mm}
\centering
{
\small
\renewcommand{\arraystretch}{0.85}
\setlength{\tabcolsep}{0.3em}
\begin{tabular}{lccccc}
\toprule
\textbf{Method} & 
Abs Rel &
Sq Rel &
RMSE &
RMSE$_{log}$ &
$\delta_{1.25}$ \\
\midrule
ResNet18 & 0.218 & 2.053 & 8.154 & 0.355 & 0.650 \\
ResNet18$^\ddagger$ & 0.212 & 1.918 & 7.958 & 0.323 & 0.674 \\
ResNet50 & 0.216 & 2.165 & 8.477 & 0.371 & 0.637 \\
ResNet50$^\ddagger$ & 0.210 & 2.017 & 8.111 & 0.328 & 0.697 \\
\textbf{PackNet}         & \textbf{0.187} & \textbf{1.852} & \textbf{7.636} & \textbf{0.289} & \textbf{0.742} \\
\bottomrule
\end{tabular}
}
\vspace{-1mm}
\caption{\textbf{Generalization capability of different depth networks}, trained on both KITTI and CityScapes and evaluated on NuScenes \cite{nuscenes}, for 640 x 192 resolution and distances up to 80m. $^\ddagger$ denotes ImageNet \cite{Deng09imagenet} pretraining. }
\label{tab:nuscenes}
\vspace{-5mm}
\end{table}


\subsection{Generalization Capability}

\textcolor{black}{
We also investigate the generalization performance of \textit{PackNet}, as evidence that it does not simply memorize training data but learns transferable discriminative features. To assess this, we evaluate on the recent NuScenes dataset~\cite{nuscenes} models trained on a combination of CityScapes and KITTI (CS+K), without any fine-tuning.
Results in Table~\ref{tab:nuscenes} show \textit{PackNet} indeed generalizes better across a large spectrum of vehicles and countries (Germany for CS+K, USA + Singapore for NuScenes), outperforming standard architectures in all considered metrics without the need for large-scale supervised pretraining on ImageNet.
}

\vspace{-2mm}
\section{Conclusion}
\label{sec:conclusion}
\textcolor{black}{
We propose a new convolutional network architecture for \emph{self-supervised} monocular depth estimation: \textit{PackNet}. It leverages novel, symmetrical, detail-preserving \textit{packing} and \textit{unpacking} blocks that jointly learn to compress and decompress high resolution visual information for fine-grained predictions. Although purely trained on unlabeled monocular videos, our approach outperforms other existing self- and semi-supervised methods and is even competitive with fully-supervised methods while able to run in real-time. It also generalizes better to different datasets and unseen environments without the need for ImageNet pretraining, especially when considering longer depth ranges, as assessed up to 200m on our new DDAD dataset. Additionally, by leveraging during training only weak velocity information, we are able to make our model scale-aware, i.e. producing metrically accurate depth maps from a single image.
}


\section*{Acknowledgments}
We would like to thank John Leonard and Wolfram Burgard for their support and insightful comments during the development of this work. 

\appendix
\begin{table*}[bp!]
\centering
{
\small
\setlength{\tabcolsep}{1.0em}
\begin{tabular}{lccccc}
\toprule
\textbf{Method} &
Supervision & 
Resolution & 
GT & 
\quad Seq. 09\quad & 
\quad Seq. 10\quad \\
\midrule
SfMLearner (Zhou et al.~\cite{zhou2017unsupervised})
& M & 416 x 128 & \checkmark & 0.021 $\pm$ 0.017 & 0.020 $\pm$ 0.015 \\
Monodepth2 (Godard et al.~\cite{monodepth2})
& M & 640 x 192 & \checkmark & 0.017 $\pm$ 0.008 & 0.015 $\pm$ 0.010 \\
DF-Net (Zou et al.~\cite{zou2018dfnet})
& M & 576 x 160 & \checkmark & 0.017 $\pm$ 0.007 & 0.015 $\pm$ 0.009 \\
Vid2Depth (Mahjourian et al.~\cite{mahjourian2018unsupervised})
& M & 416 x 128 & \checkmark & 0.013 $\pm$ 0.010 & 0.012 $\pm$ 0.011 \\
GeoNet (Yin et al.~\cite{yin2018geonet})
& M & 416 x 128 & \checkmark & 0.012 $\pm$ 0.007 & 0.012 $\pm$ 0.009 \\
Struct2Depth (Casser et al.~\cite{casser2018depth})
& M & 416 x 128 & \checkmark & 0.011 $\pm$ 0.006 & 0.011 $\pm$ 0.010 \\
TwoStreamNet (Ambrus et al. \cite{ambrus2019stream})
& M & 640 x 192 & \checkmark & \textbf{0.010} $\pm$ \textbf{0.002} & \textbf{0.009} $\pm$ \textbf{0.002} \\
\midrule
\textbf{PackNet-SfM}                                             & M    & 640 x 192 & \checkmark & 0.011 $\pm$ 0.006 & \textbf{0.009} $\pm$ 0.007 \\
\textbf{PackNet-SfM}                                             & M+v  & 640 x 192 & \checkmark & \textbf{0.010} $\pm$ 0.005 & \textbf{0.009} $\pm$ 0.008 \\
\textbf{PackNet-SfM}                                             & M+v  & 640 x 192 &            & 0.014 $\pm$ 0.007  & 0.012 $\pm$ 0.008 \\
\bottomrule
\end{tabular}\\\vspace{1mm}
}
\caption{\textbf{Average Absolute Trajectory Error (ATE) in meters on the KITTI Odometry Benchmark~\cite{geiger2013vision}}: All methods are trained on Sequences 00-08 and evaluated on Sequences 09-10. The ATE numbers are averaged over all overlapping 5-frame snippets in the test sequences. \text{M+v} refers to velocity supervision (v) in addition to monocular images (M). The \textit{GT} checkmark indicates the use of ground-truth translation to scale the estimates at test-time.}
\label{table:supplementary-pose-ate}
\end{table*}

\section{Pose evaluation}

In Table~\ref{table:supplementary-pose-ate} we show the results of our proposed \textit{PackNet-SfM} framework on the KITTI odometry benchmark~\cite{geiger2013vision}. To compare with related methods, we train our framework from scratch on sequences 00-08 of the KITTI odometry benchmark, with exactly the same parameters and networks used for depth evaluation (Table 3, main text). For consistency with related methods, we compute the Absolute Trajectory Error (ATE) averaged over all  5-frame snippets on sequences 09 and 10. Note that our pose network only takes two frames as input, and outputs a single transformation between that pair of frames. To evaluate our model on 5-frame snippets we combine the relative transformations between the target frame and the first context frame into 5-frame long overlapping trajectories, stacking $f_x\left(I_t,I_{t-1}\right) = x_{t\to t-1}$ to create appropriately sized trajectories. 

\begin{figure*}[t!]
    \centering
    \includegraphics[width=0.32\textwidth]{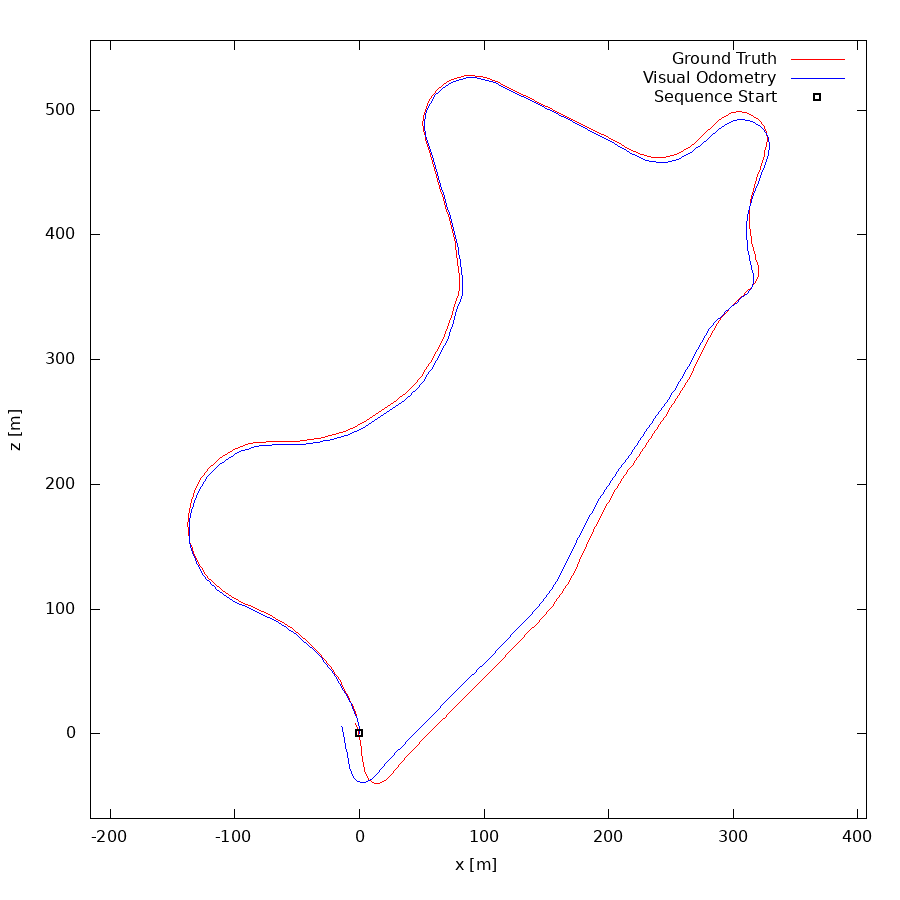}
    \includegraphics[width=0.32\textwidth]{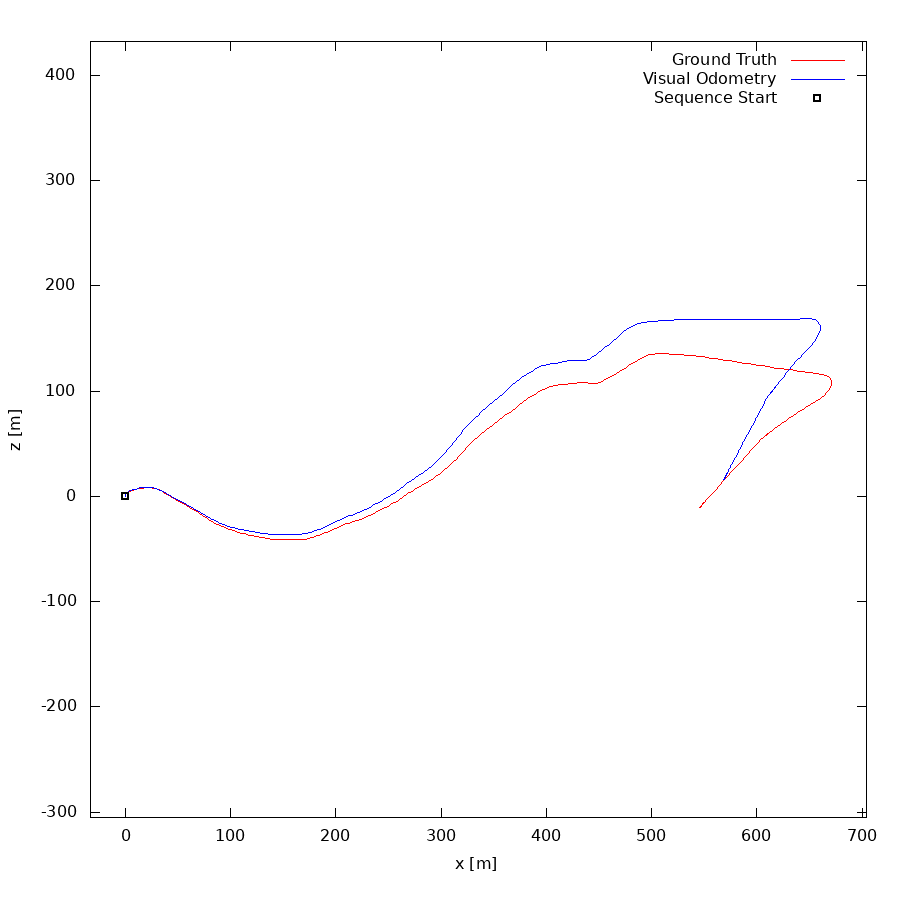}
    \caption{\textbf{Pose evaluation on KITTI test sequences.} Qualitative trajectory results of \textit{PackNet-SfM} on test sequences 09 and 10 of the KITTI odometry benchmark.}
    \label{fig:supplementary-pose-trajectory-test}
\vspace{-3mm}
\end{figure*}

The ATE results are summarized in Table~\ref{table:supplementary-pose-ate}, with our proposed framework achieving competitive results relative to other related methods. We also note that all these related methods are trained in the monocular setting (M), and therefore scaled at test-time using ground truth information. Our method, on the other hand, when trained with the proposed velocity supervision loss (M+v) does not require ground-truth scaling at test-time, as it is able to recover metrically accurate scale purely from monocular imagery. Nevertheless, it is still able to achieve competitive results compared to other methods. Examples of reconstructed trajectories obtained using \emph{PackNet-SfM} for the test sequences can be found in Figure \ref{fig:supplementary-pose-trajectory-test}.

\section{Dense Depth for Automated Driving (DDAD)}

\begin{figure*}[b!]
    \centering
    \includegraphics[width=0.95\textwidth]{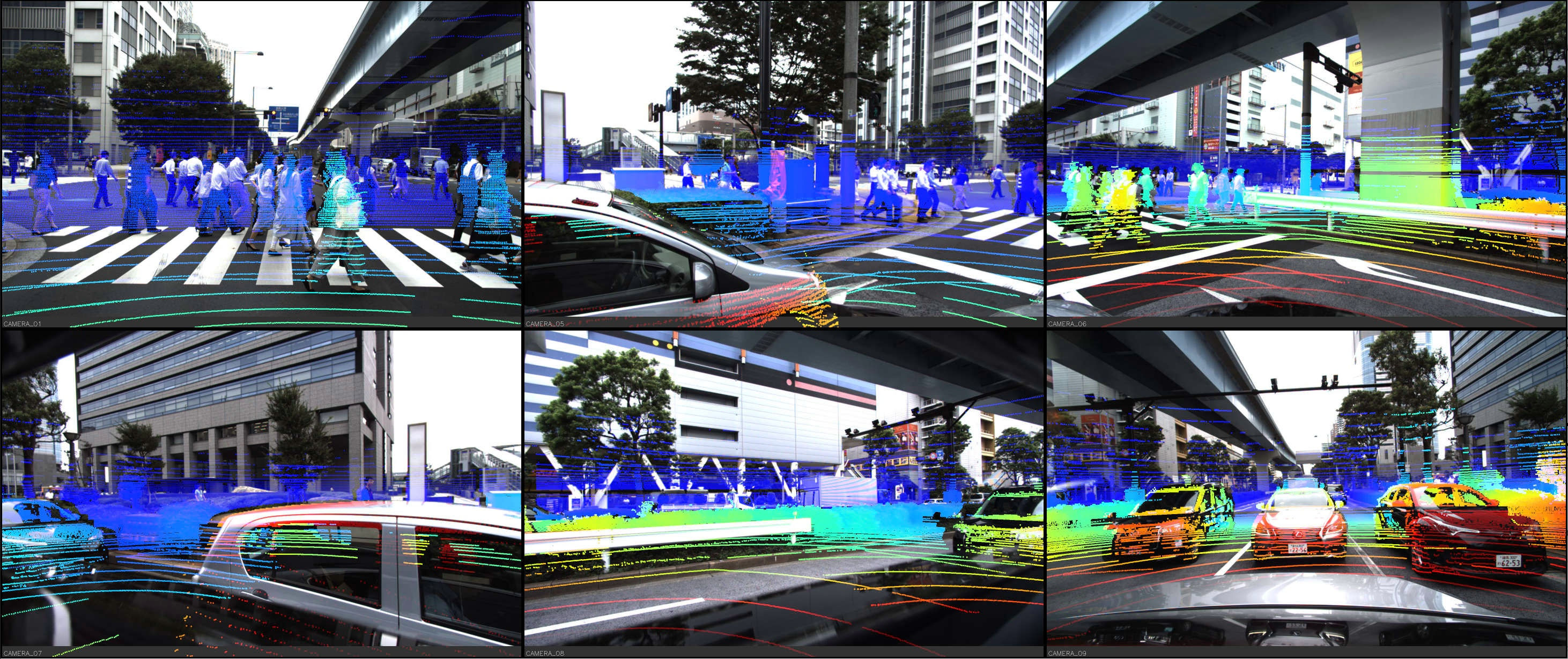}
    \caption{\textbf{DDAD sample} from Tokyo, Japan.}
    \label{fig:supplementary-ddad-viz1}
\end{figure*}

\begin{figure*}[t!]
    \centering
    \includegraphics[width=0.98\textwidth]{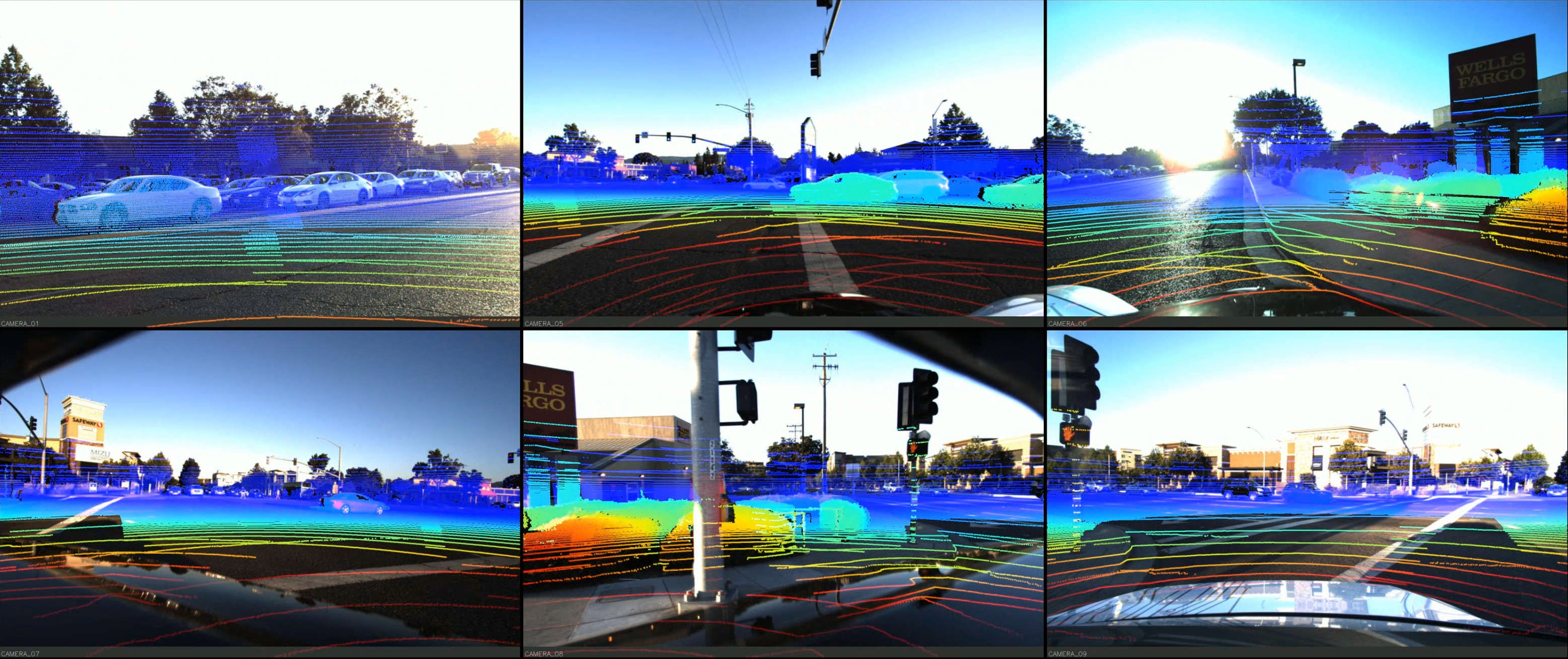}
    \caption{\textbf{DDAD sample} from San Francisco Bay Area, California.}
    \label{fig:supplementary-ddad-viz2}
\end{figure*}

\begin{figure*}[t!]
    \centering
    \includegraphics[width=0.98\textwidth]{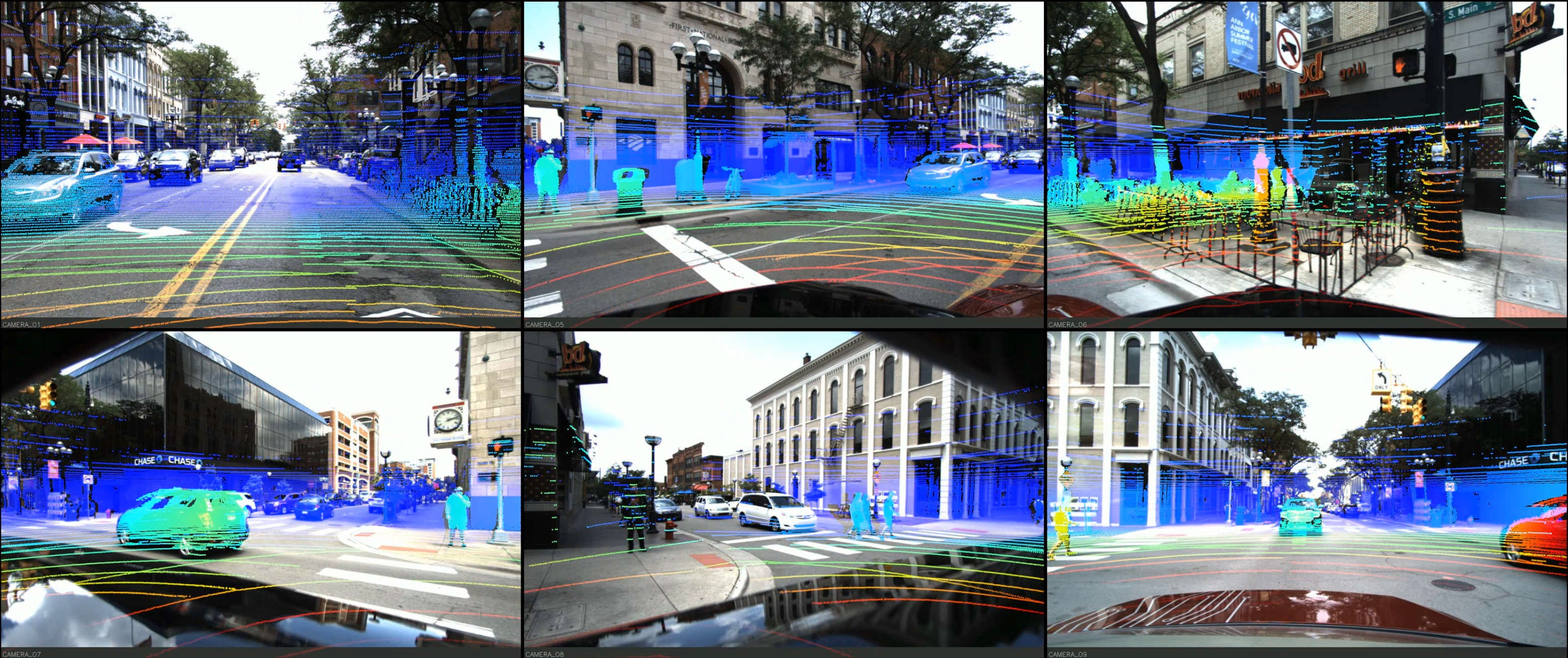}
    \caption{\textbf{DDAD sample} from Detroit, Michigan.}
    \label{fig:supplementary-ddad-viz3}
\end{figure*}

In this section, we provide a brief overview of our newly introduced \emph{DDAD (Dense Depth for Automated Driving)} dataset and the relevant properties that make it desirable as a dense depth estimation benchmark. 
It includes a high-resolution, long-range Luminar-H2\footnote{https://www.luminartech.com/technology} as the LiDAR sensor used to generate pointclouds, with a maximum range of 250m and sub-1cm range precision. 
Additionally, it contains six calibrated cameras time-synchronized at 10 Hz, that together produce a 360$^\circ$ coverage around the vehicle. Note that in our work we only use information from the front-facing camera for training and evaluation.

Examples of a Luminar-H2 pointcloud projected onto each of these six cameras are shown in Figures~\ref{fig:supplementary-ddad-viz1}, ~\ref{fig:supplementary-ddad-viz2} and ~\ref{fig:supplementary-ddad-viz3}, for different urban settings. 
The depth maps generated from projecting these Luminar pointclouds onto the camera frame allow us to evaluate depth estimation methods in a much more challenging way, both in terms of denseness and longer ranges. In Table 2 and Figure 6 of the main text we show how our proposed \emph{PackNet} architecture outperforms other related methods under these conditions. In fact, the gap in performance increases when considering denser ground-truth information at longer ranges, both on the entire interval and at discretized bins.

DDAD is a cross-continental dataset with scenes drawn from urban settings in the United States (San Francisco Bay Area, Detroit and Ann Arbor) and Japan (Tokyo and Odaiba). Each scene is 5 or 10 seconds long and consists of 50 or 100 samples with corresponding Luminar-H2 pointcloud and six image frames, including intrinsic and extrinsic calibration. The training set contains 194 scenes with a total of 17050 individual samples, and the validation set contains 60 senes with a total of 4150 samples. \textcolor{black}{The six cameras are $2.4$ MP ($1936 \times 1216$), global-shutter, and oriented at $60\degree$ intervals. They are synchronized with $10$ Hz scans from our Luminar-H2 sensors oriented at $90\degree$ intervals.}
















{\small
\bibliographystyle{ieee_fullname}
\bibliography{main}
}

\end{document}